\documentclass[11pt]{article}

\usepackage[final]{acl}

\usepackage{times}
\usepackage{latexsym}
\usepackage[frozencache, cachedir=minted-cache]{minted}
\usepackage{caption}
\usepackage{booktabs}
\usepackage{multirow}
\usepackage{titlesec}
\usepackage{amssymb} 
\usepackage{pifont}
\usepackage{xcolor}

\newcommand{\iconSize}{\small}

\newcommand{\xmark}{{\iconSize\textcolor{red!80!black}{\ding{55}}}}

\usepackage{xcolor}
\usepackage{pifont}   

\DeclareRobustCommand{\gcir}{{\iconSize\textcolor{green!70!black}{\ding{108}}}}   
\DeclareRobustCommand{\ycir}{{\iconSize\textcolor{yellow!60!black}{\ding{108}}}}  
\DeclareRobustCommand{\ocir}{{\iconSize\textcolor{orange!90!black}{\ding{108}}}}  
\DeclareRobustCommand{\rcir}{{\iconSize\textcolor{red!75!black}{\ding{108}}}}     

\usepackage[T1]{fontenc}

\usepackage[utf8]{inputenc}

\usepackage{microtype}

\usepackage{inconsolata}

\usepackage{graphicx}

\usepackage{array}
\newcolumntype{L}[1]{>{\raggedright\let\newline\\\arraybackslash\hspace{0pt}}m{#1}}
\newcolumntype{C}[1]{>{\centering\let\newline\\\arraybackslash\hspace{0pt}}m{#1}}
\newcolumntype{R}[1]{>{\raggedleft\let\newline\\\arraybackslash\hspace{0pt}}m{#1}}
\title{\textsc{EduAdapt}: A Question Answer Benchmark Dataset for Evaluating Grade-Level Adaptability in LLMs}

\author{\textbf{Numaan Naeem$^{\delta}$} \quad  
        \textbf{Abdellah El Mekki$^{\gamma}$}\quad 
        \textbf{Muhammad Abdul-Mageed$^{\gamma}$} \quad
     \\
$^{\delta}$MBZUAI \quad ~~~~
$^{\gamma}$The University of British Columbia \quad \\
\texttt{numaan.naeem@mbzuai.ac.ae} \\
\texttt{\{abdellah.elmekki, muhammad.mageed\}@ubc.ca}
}

\begin{document}
\maketitle
\begin{abstract}
Large language models (LLMs) are transforming education by answering questions, explaining complex concepts, and generating content across a wide range of subjects. Despite strong performance on academic benchmarks, they often fail to tailor responses to students’ grade levels. This is a critical need in K-12 education, where age-appropriate vocabulary and explanation are essential for effective learning. Existing models frequently produce outputs that are too advanced or vague for younger learners, and there are no standardized benchmarks to evaluate their ability to adjust across cognitive and developmental stages. To address this gap, we introduce \textsc{EduAdapt}, a benchmark of nearly 48k grade-labeled QA pairs across nine science subjects, spanning Grades 1-12 and grouped into four grade levels. We evaluate a diverse set of open-source LLMs on \textsc{EduAdapt} and find that while larger models generally perform better, they still struggle with generating suitable responses for early-grade students (Grades 1-5). Our work presents the first dataset and evaluation framework for assessing grade-level adaptability in LLMs, aiming to foster more developmentally aligned educational AI systems through better training and prompting strategies. \textsc{EduAdapt} code and datasets are publicly available.\footnote{ \texttt{\href{https://github.com/NaumanNaeem/EduAdapt}{https://github.com/NaumanNaeem/EduAdapt}}}
\end{abstract}

\section{Introduction}

Recent research has shown that LLMs can perform at a student level on standardized tests across subjects like mathematics, physics, and computer science, often achieving high accuracy on both multiple-choice and open-ended questions~\cite{openai2024gpt4technicalreport}. For example, studies demonstrate that tools like ChatGPT are capable of generating logically coherent responses that reflect a strong grasp of subject matter across a wide range of disciplines ~\cite{susnjak2022chatgptendonlineexam}. While these abilities are impressive, they mainly benefit older students. Prior work shows that LLMs often fail to adapt explanations to grade levels, producing responses too complex for younger learners or oversimplified for advanced ones, and even when prompted, they struggle to adjust language, tone, and complexity to different age groups~\cite{rooein2023knowaudiencellmsadapt}.

\begin{figure}[t]
  \centering
  \includegraphics[width=1\linewidth]{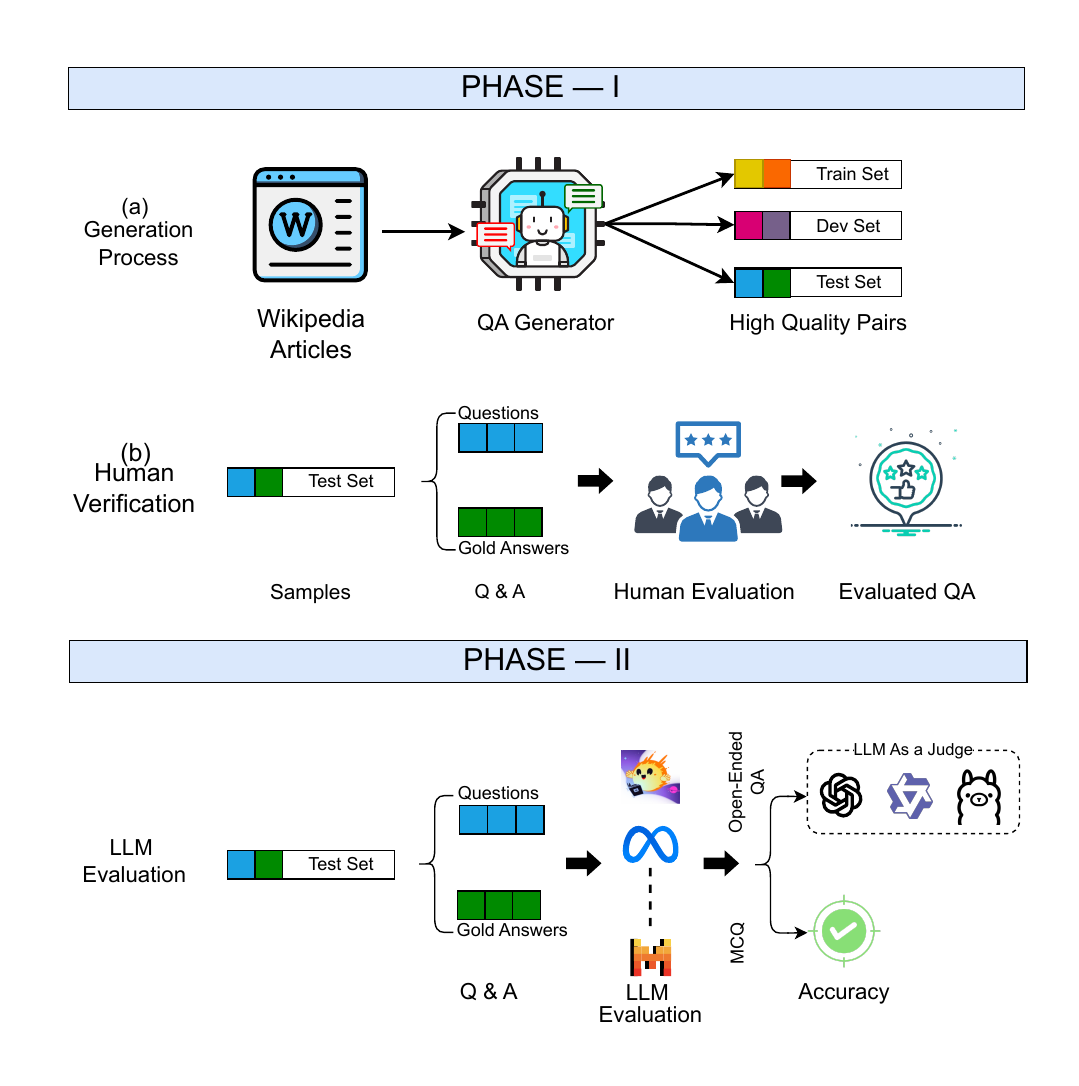}
  \caption{Overview of the whole pipeline. \textbf{Phase~I }covers the \textsc{EduAdapt} creation process, which includes Generation and Human Verification, where Wikipedia articles are transformed into QA pairs and manually validated for quality. \textbf{Phase~II} covers LLM Evaluation, where open-source LLMs are tested on \textsc{EduAdapt}'s Test set to assess grade-level adaptability.}
  \label{fig:main_fig}
\end{figure}

This is particularly concerning, given the high level of digital engagement among children. According to UNICEF, one in three internet users globally is a child \cite{keeley2017state}, and children aged 8-12 spend over five hours per day on screens on average \cite{rideout2022commonsense}. This presents a major opportunity to enrich learning through AI, but also a risk if content is not age-appropriate or understandable. The key concerns include a lack of contextual relevance for younger users \cite{nayeem2024kidlm, seo2024chacha} and difficulties in maintaining the right level of lexical simplicity across grade levels \cite{valentini2023automatic}.

To address these challenges, researchers have developed specialized models such as KidLM \cite{nayeem2024kidlm}, trained on child-appropriate data with objectives aimed at improving readability, safety, and developmental suitability. Such domain-specific efforts highlight the need for LLMs that are not only accurate but also adaptive to the diverse educational needs of younger audiences. This paper tackles a key challenge towards these LLMs: the current inability of LLMs to effectively adjust their responses across grade levels.

To bridge this gap, we developed \textsc{EduAdapt}, a benchmark of nearly 48k QA pairs (comprising both multiple-choice and open-ended question-answers) for Grades 1-12 across nine educational subjects, following the K-12 framework. To capture developmental progression, we reorganized grades into four levels (Grades 1-2, 3-5, 6-8, and 9-12). All QA pairs are aligned with the Next Generation Science Standards (NGSS) \cite{ngss2013}, ensuring coverage from basic recall to higher-order reasoning. \textsc{EduAdapt} enables both adaptation (through the released training set) and evaluation (through the released test set) of LLM knowledge at specific grade levels. In our experiments, we evaluated existing open-source LLMs across different sizes and families on the \textsc{EduAdapt} test set. The results show that even leading LLMs struggle to adjust their outputs consistently across grade levels, underscoring the need for targeted educational LLMs. To the best of our knowledge, \textsc{EduAdapt} is the first benchmark designed specifically for evaluating grade-level adaptability of LLMs across the full K-12 system.

\section{Methodology}
\label{sec:methodology}

The creation of the \textsc{EduAdapt} Benchmark dataset involves two main stages: the \textbf{Generation Process}, where we automatically generate the QA benchmark, and the \textbf{Human Verification Process}, where we verify and revise the generated data to make sure it is of high quality. The \textbf{Generation Process} begins with extracting clean, domain-specific text from Wikipedia articles and creating seed QA pairs tailored to different educational levels using a reliable LLM. These pairs are then refined through a self-reflection mechanism \cite{renze2024self}, enabling the LLM to evaluate and improve its outputs based on pedagogical criteria, ultimately filtering high-quality pairs for the final dataset. The \textbf{Human Verification} ensures that human reviewers assess a subset of the dataset for quality and grade-level appropriateness, validating its educational soundness. Phase I in Figure~\ref{fig:main_fig} provides an overview of the complete methodology, summarizing the process from data collection to human verification.

\subsection{Stage 1: Generation Process}

\begin{figure*}[h]
  \centering
  \includegraphics[width=0.9\linewidth]{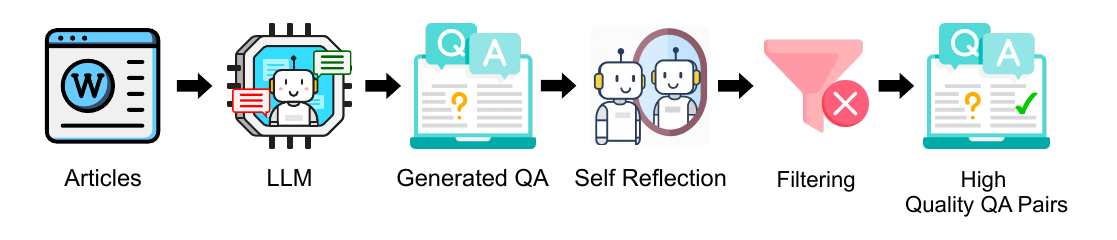}
  \caption{\textbf{QA Generator:} A pipeline that generates QA pairs. Wikipedia passages are first fed into the QA generator to produce QA pairs. Each pair is then evaluated by a self-reflection module, which scores it across five defined criteria. Filtering is applied to retain only those pairs with ratings above 8 on all criteria, resulting in a high-quality QA dataset.}
  \label{fig:generator}
\end{figure*}

The first stage of our pipeline involves extracting and cleaning Wikipedia articles, which serve as input for the question-answer generation process, as shown in Phase I (part a) of Figure~\ref{fig:main_fig}. Each article is processed by the QA Generator, as shown in Figure~\ref{fig:generator}, to produce grade-appropriate educational QA pairs. This is implemented using the text-generation module\footnote{\url{https://distilabel.argilla.io/1.5.3/components-gallery/tasks/textgeneration/}} of the Distilabel framework \cite{distilabel2025}, which supports iterative refinement through AI-generated feedback, enhancing both data quality and model behavior.

\subsubsection{Content Collection from Wikipedia}

We collected source material from Wikipedia dumps across nine subjects: (1) \textbf{Chemistry}, (2) \textbf{Computer Science}, (3) \textbf{Meteorology}, (4) \textbf{Ecology}, (5) \textbf{Geology}, (6) \textbf{Biology}, (7) \textbf{Physics}, (8) \textbf{Medicine}, and (9) \textbf{Geography}. These subjects were selected to provide broad coverage of scientific and technical domains, ensuring a versatile dataset for educational use. The raw articles were processed through a structured pipeline that removed non-textual elements (e.g., images, tables, references), discarded irrelevant sections (e.g., biographies, timelines, historical overviews), and stripped hyperlinks and markup to obtain clean text.  

Wikipedia content, however, tends to be written at higher reading levels. Prior studies confirm this: \citet{Lucassen_Dijkstra_Schraagen_2012} found that 75\% of Wikipedia articles fall below standard readability thresholds, and \citet{wang2020assessing} reported an average reading grade of 12.7 across medical entries. Our classification results corroborate these findings. Using \textbf{Phi-4} \cite{abdin2024phi4technicalreport}, a 14B parameter model optimized for educational and reasoning tasks, we categorized passages by grade level. As shown in Table~\ref{tab:wiki_articles}, the vast majority of passages were labeled as suitable for Grades~6-12, with only about 14\% judged appropriate for Grades~1-5. This distribution underscores the challenge of sourcing age-appropriate material from Wikipedia and validates the concerns raised in prior work. This classification step yielded passages tailored to distinct grade levels, and we used these grade-appropriate texts to generate QA pairs aligned with their corresponding levels. The classification prompt is provided in Listing~\ref{listing:classify} in Appendix~\ref{subsec:appendixA.1}.

\setlength{\tabcolsep}{18pt}
\begin{table*}[!t]
\renewcommand{\arraystretch}{1.5}
\scriptsize
\centering
\begin{tabular}{l c c c c c}
    \toprule
    \textbf{Subject} & \textbf{Grade 1--2} & \textbf{Grade 3--5} & \textbf{Grade 6--8} & \textbf{Grade 9--12} & \textbf{Total Passages}\\
    \midrule
    Biology           & 1,211 & 686  & 2,443 & 9,349  & 13,689 \\
    Physics           & 558  & 267  & 1,070 & 12,756 & 14,651 \\
    Chemistry         & 532  & 382  & 1,241 & 11,452 & 13,607 \\
    Computer Science  & 1,000 & 451  & 2,080 & 9,661  & 13,192 \\
    Ecology           & 1,797 & 1,151 & 3,997 & 6,681  & 13,626 \\
    Geography         & 3,487 & 1,193 & 2,975 & 5,336  & 12,991 \\
    Geology           & 1,284 & 627  & 2,331 & 8,730  & 12,972 \\
    Medicine          & 1,024 & 458  & 1,807 & 10,663 & 13,952 \\
    Metrology         & 692  & 367  & 1,576 & 10,690 & 13,325 \\
    \midrule
    \textbf{Total}    & 11,585 & 5,582 & 17,891 & 85,318 & 122,005 \\
    \bottomrule
\end{tabular}
\caption{Distribution of Wikipedia passages by subject and grade level after classification with \textbf{Phi-4}. Only passages identified as suitable for Grades~1-5 were retained for early-grade QA generation.}
\label{tab:wiki_articles}
\end{table*}

\subsubsection{Question-Answer Generation}

After extracting and cleaning domain and grade-specific texts from Wikipedia, we used this curated content to generate QA pairs for each grade group of K-12, encompassing both multiple-choice questions (MCQs) and open-ended QA. These pairs were designed to reflect cognitive abilities across the K-12 spectrum. Prior work has shown the importance of combining these two formats: \citet{lin2022truthfulqameasuringmodelsmimic} designed \textsc{TruthfulQA} to assess factual accuracy through both fixed-choice and free-form answers, while \citet{yue2025mmmuprorobustmultidisciplinemultimodal} introduced \textsc{MMMU}, a large-scale multimodal benchmark that blends mostly MCQs with some open-ended tasks to evaluate deeper reasoning. Following the same approach, we include both MCQs and open-ended questions to provide a more comprehensive evaluation of model performance.

To ensure that QA pairs matched students’ language and cognitive abilities, we designed tailored prompts for each grade level, drawing on NGSS guidelines \cite{ngss2013} to reflect appropriate comprehension and reasoning skills. These prompts underwent iterative refinement through pilot runs and manual inspection, where unclear wording or overly advanced phrasing was adjusted to improve grade level adaptability. Once finalized, they were integrated into a structured QA generation pipeline.  The finalized prompts are shown in Listings~\ref{listing:grade1_2} through~\ref{listing:grade9_12} in Appendix ~\ref{subsec:appendixA.2}. This stage produced approximately \textbf{166k} QA pairs (consisting of both multiple-choice and open-ended question-answers) across all subjects and grade levels.

\subsubsection{Self-Reflection for Quality Assessment}

Self-reflection is a technique in which an LLM critiques its own output to improve reliability and alignment \cite{renze2024self}. It involves feeding the model's own output back to itself as part of a new prompt. This new prompt explicitly asks the model to evaluate the previous response on defined criteria. The goal is to simulate a human-like review process where the model identifies potential errors or areas for improvement.

In our work, after generating QA pairs from the previous section, we applied this mechanism using the same model to evaluate its own outputs against pedagogical and linguistic standards informed by NGSS guidelines.\footnote{Prior work shows that the same model can effectively serve both generation and evaluation \cite{renze2024self}.}  We implemented a customized UltraFeedback-style \footnote{\url{https://distilabel.argilla.io/1.5.3/components-gallery/tasks/ultrafeedback/}} pipeline \cite{cui2024ultrafeedbackboostinglanguagemodels} with the Distilabel framework, adapting it for educational QA. Evaluation focused on five evaluation criteria: language appropriateness, grade alignment, relevance, clarity, and subject fit, each tailored to the developmental stage of the target grade level. Every QA pair received a 1-10 score on each criteria, with the average serving as the overall rating. To ensure consistency, we retained only pairs scoring at least 8 on every criterion. From the initial \textbf{166k} pairs, only \textbf{47,734} passed this filtering step.

Our final dataset contains \textbf{47,734} QA pairs, divided into \textbf{28,640} for training, \textbf{9,547} for development, and \textbf{9,547} for testing. It includes a mix of open-ended and multiple-choice questions, ensuring broad coverage of student proficiency levels while supporting both benchmarking and fine-tuning. Overall and per-subject distributions are reported in Table~\ref{table:data}, with detailed splits by subset shown in Tables~\ref{table:train_data}, \ref{table:dev_data}, and \ref{table:test_data} in Appendix~\ref {sec:AppendixB}. The prompts used for self-reflection are provided in Listings~\ref{listing:grade_1_2_self-reflection_prompt}--\ref{listing:grade_9_12_self-reflection_prompt} in Appendix~\ref{subsec:appendixA.3}.

\paragraph{Error Analysis of Low-Quality QA Pairs.} Although the final dataset retains only QA pairs scoring above 8 on every criterion, the pairs that fell short still offer valuable insights. We analyzed QA pairs with an \textit{average self-reflection rating} below 8 to identify recurring sources of error. To structure the analysis, pairs were divided into two groups:  
\textbf{(1)} those with an average rating below 5, and  
\textbf{(2)} those with ratings between 5 and 8.  
From each grade level and subject, we randomly sampled five pairs per group, yielding a total of 360 QA pairs. A summary of the evaluation results is provided in Table~\ref{tab:analysis_1}. Pairs with ratings below 5 mainly failed because of poor grade alignment. Many used concepts or terms that were too advanced for the target students. At the lower grades (1-5), some questions were oversimplified to the point of being off-topic, while at higher grades (6-12), they often focused on very technical or uncommon knowledge that was not suitable.

Pairs rated between 5 and 8 showed different problems. For lower grades, the language was clearer, but the questions often drifted away from core subject knowledge, focusing more on everyday details than scientific ideas. For higher grades, the questions tended to be slightly too complex, which caused relevance issues. Overall, in both groups, the model usually produced questions that were clear but often failed to capture the real subject knowledge for the appropriate grade level. This shows that all evaluation criteria, language appropriateness, grade alignment, relevance, clarity, and subject fit need to be considered equally to ensure high-quality QA pairs.

\setlength{\tabcolsep}{3pt}
\begin{table*}[]
\renewcommand{\arraystretch}{1.4}
\scriptsize
\centering
\begin{tabular}{ l| c| c  c  c  c  c c c  c c} 
\toprule
    {} & \textbf{Grade Levels} & \textbf{Biology} & \textbf{Physics} & \textbf{Chemistry} & \textbf{Computer Science} & \textbf{Ecology} & \textbf{Geography} & \textbf{Geology} & \textbf{Medicine} & \textbf{Metrology} \\
    \midrule
    & 1 and 2 & 462 & 80 & 125 & 154 & 790  & 1220 & 256 & 146 & 92  \\
    \multirow{-9}{*}{\rotatebox[origin=c]{90}{\parbox[c]{4cm}{\texttt{\textbf{QA Count}~~~}}}} &   3 to 5  & 1004 & 100 & 263 & 438 & 1510  & 2086 & 236 & 379 & 124 \\ 
    & 6 to 8 & 344 & 89 & 144 & 302 & 759  & 463 & 299 & 275 & 221 \\
    & 9 to 12 & 1407 & 2475 & 2706 & 2263 & 1159 & 687 & 1248 & 1621 & 1660  \\
    \midrule
    & 1 and 2 & 191 & 44 & 104 & 48 & 220 & 229 & 40 & 53 & 37 \\
    \multirow{-10}{*}{\rotatebox[origin=c]{90}{\parbox[c]{4.5cm}{\texttt{\textbf{MCQ Count}~~~}}}} &   3 to 5  & 363 & 43 & 82 & 111 & 736 & 913 & 50 & 85 & 100  \\ 
    & 6 to 8 & 409 & 95 & 150 & 311 & 925 & 463 & 417 & 187 & 118  \\
    & 9 to 12 & 1125 & 1972 & 2421 & 1983 & 1105 & 592 & 1323 & 1251 & 1851 \\
    
\midrule 
    \rotatebox[origin=c]{90}{\parbox[c]{0.5cm}{\texttt{\textbf{Total}}}} &   {}  & \textbf{5305} & \textbf{4898} & \textbf{5995} & \textbf{5610} & \textbf{7204} & \textbf{6653} & \textbf{3869} & \textbf{3997} & \textbf{4203} \\ 
\bottomrule

\end{tabular}

\caption{\textbf{Full Dataset}: Distribution of question-answer pairs across all subjects and grade levels in the full dataset.
}
\label{table:data}
\end{table*}

\setlength{\tabcolsep}{5pt}
\begin{table*}[]
    \centering
    \renewcommand{\arraystretch}{1.55}
    \scriptsize
    \begin{tabular}{l c c c c c c c}
    \toprule
    \midrule
     \textbf{Grade Level} & \textbf{Ratings} & \textbf{Pairs} & \textbf{Language Appropriateness} & \textbf{Grade Alignment} & \textbf{Relevance} & \textbf{Clarity} & \textbf{Subject-Fit} \\
     \midrule 
     \midrule
     Grade 1--2   & $1 \leq r < 5$  & 45 & 5.40 & 3.02 & 2.37 & 6.80 & 2.80 \\
Grade 3--5   & $1 \leq r < 5$  & 45 & 5.29 & 3.67 & 2.95 & 6.73 & 3.27 \\
Grade 6--8   & $1 \leq r < 5$  & 45 & 5.18 & 3.56 & 3.16 & 6.07 & 3.09 \\
Grade 9--12  & $1 \leq r < 5$  & 45 & 5.25 & 3.56 & 3.06 & 6.68 & 2.69 \\

     \midrule
Grade 1--2   & $5 \leq r \leq 8$  & 45 & 7.92 & 6.17 & 5.70 & 8.31 & 6.36 \\
Grade 3--5   & $5 \leq r \leq 8$  & 45 & 7.71 & 6.67 & 6.37 & 8.22 & 6.71 \\
Grade 6--8   & $5 \leq r \leq 8$  & 45 & 7.31 & 6.22 & 5.96 & 7.62 & 6.44 \\
Grade 9--12  & $5 \leq r \leq 8$  & 45 & 7.78 & 6.38 & 5.93 & 8.18 & 5.96 \\

\bottomrule
\end{tabular}
\caption{Evaluation results for QA pairs excluded during filtering (\(r < 8\)). Each row shows the average scores across 45 sampled pairs per grade level and rating band. Lower ratings highlight recurring weaknesses in grade alignment, relevance, and subject fit.}

\label{tab:analysis_1}
\end{table*}

\subsection{Stage 2: Human Verification Process}

In the second stage, we assessed the quality of the generated dataset through human evaluation of approximately 10\% of the test set (1,000 QA pairs). We sampled \textbf{28} pairs per grade per subject, except for metrology and physics, where \textbf{20} pairs were reviewed in Grades 1-5 and \textbf{34} in Grades 6-12.  

We hired three reviewers with backgrounds in educational content development via the Upwork platform to independently evaluate the QA pairs. Each annotator was compensated at a rate of \$4 per 100 words for reviewing the complete set of 1k QA pairs. Using the same criteria as in Listings~\ref{listing:grade_1_2_self-reflection_prompt} to \ref{listing:grade_9_12_self-reflection_prompt} in Appendix~\ref{subsec:appendixA.3}, each pair was scored from 1 to 10 on language appropriateness, grade alignment, relevance, clarity, and subject fit. An overall score was then calculated as the average across criteria. Table~\ref{tab:human_evaluation_stats} reports the average scores assigned by each reviewer, while Table~\ref{tab:avg_across_human} in Appendix~\ref{sec:AppendixB} presents aggregated results across criteria and grade levels for all humans. Together, these highlight both reviewer consistency and grade-level trends in dataset quality.

\begin{table}[!h]
    \centering
    \scalebox{0.75}{
    \begin{tabular}{l c c c c}
    \toprule
    \midrule
     \textbf{Grade Level} & \textbf{Human 1} & \textbf{Human 2} & \textbf{Human 3} & \textbf{Average}\\
     \midrule \midrule
     Grade 1 and 2       & 7.18 & 7.71 & 8.19 & 7.69\\ 
     Grade 3 to 5        & 8.14 & 7.56 & 8.32 & 8.00\\
     Grade 6 to 8        & 8.20 & 7.58 & 8.63 & 8.14\\
     Grade 9 to 12       & 9.00 & 8.86 & 8.71 & 8.86\\ \hline
    \end{tabular}
    }
    \caption{Average human evaluation scores on a scale of 1 to 10 across grade levels on the sample of the test set.}
    \label{tab:human_evaluation_stats}
\end{table}

The human evaluation results offer a comprehensive view of dataset quality from an expert perspective. To assess reliability, we measured inter-annotator agreement using Fleiss’ Kappa \cite{fleiss1971measuring} across the three reviewers, computing scores per grade level by averaging agreement across all nine subjects. Table~\ref{tab:human_kappa_scores} reports the results, which show consistently high agreement, reinforcing the credibility of the human ratings as a benchmark for dataset quality. Overall, \textit{the strong inter-rater consistency confirms the dataset’s reliability} while pointing to areas for future refinement.

\section{Experiments}
As a testbed of \textsc{EduAdapt} test set, we benchmark a group of leading existing open-source instructed-tuned LLMs to evaluate the capabilities in answering questions for different grade levels.

\paragraph{Evaluated Models.} We assessed model performance with a focus on grade-level appropriateness. To capture the effects of model architecture and scale, we evaluated a diverse set of language models, including Qwen2.5 models at 1.5B, 3B, 7B, and 14B parameters~\cite{qwen2.5}, SmolLM-1.7B~\cite{allal2024SmolLM}, Gemma-2B-it~\cite{gemmateam2024gemma2improvingopen}, LLaMA3.2-3B and LLaMA3-8B~\cite{meta2024llama3.2, llama3modelcard}, and Mistral-Small-24B~\cite{mistral2025small3}. Table~\ref{tab:model_table} in the Appendix \ref{sec:AppendixB} summarizes the Hugging Face identifiers and roles of each model used in our pipeline. By evaluating a diverse set of models, we analyzed how differences in architecture and size influence the ability to produce accurate, grade-aligned, and pedagogically sound responses. This also revealed trade-offs between model size, computational cost, and response quality in educational settings.  Each model was evaluated using the same set of questions, with prompts explicitly tailored to indicate the intended grade level of the target learners. The prompt used during evaluation for response generation is shown in Listing~\ref{listing:mode_evaluation_prompt} in Appendix~\ref{subsec:appendixA.5}. 

\paragraph{Evaluated Method.} For open-ended questions, models received only the question text and were expected to generate a grade-appropriate answer. For MCQs, the full question with answer options was provided, and models had to select the correct choice.

\paragraph{Performance Metrics.} MCQ performance was measured using accuracy, while open-ended responses were assessed through an LLM-as-a-judge framework. Three independent judges, \textbf{Qwen2.5-72B} \cite{qwen2025qwen25technicalreport}, \textbf{LLaMA3.3-70B} \cite{grattafiori2024llama3herdmodels}, and \textbf{GPT-4o} \cite{openai2024gpt4technicalreport}, scored each response against the reference answer on a 1-10 scale across 5 qualitative criteria. The final score was computed as the average of these ratings. The evaluation setup is illustrated in Figure~\ref{fig:main_fig} (Phase II), and the full judging prompt is shown in Listing~\ref{listing:judge_prompt} in Appendix~\ref{subsec:appendixA.4}.

\begin{table}[!h]
    \centering
    \scalebox{1}{
    \begin{tabular}{l c }
    \toprule
    \midrule
     \textbf{Grade Level} & \textbf{Fleiss’ Kappa}\\
     \midrule 
     \midrule
     Grade 1 and 2       & 0.668 \\ 
     Grade 3 to 5        & 0.706 \\
     Grade 6 to 8        & 0.741\\
     Grade 9 to 12       & 0.860\\ \hline
    \end{tabular}
    }
    \caption{Average Fleiss’ Kappa scores across grade levels, over all subjects.}
    \label{tab:human_kappa_scores}
\end{table}

\paragraph{Implementation Details.}
All experiments were conducted on NVIDIA RTX A6000 GPUs (48GB). Large models such as Mistral-Small-24B, Qwen2.5-72B, and LLaMA3.3-70B were run on two GPUs, while others used a single GPU. Models were served with vLLM \cite{kwon2023efficientmemorymanagementlarge} for efficient large-scale inference. Across both generation and evaluation, we found that using a temperature of 0.3 with a top\_p value of 0.9 produced the most stable and reliable outputs.

\section{Results and Analysis}

In this section, we report our experimental results, revealing the strengths and limitations of current open-source LLMs in answering educational questions across different grade levels.

\paragraph{Results on MCQs set.} 
The model-wise average accuracy for MCQs across grade levels is reported in Table~\ref{tab:mcq_accuracy_compact}. Larger models such as Qwen2.5-14B and Mistral-Small-24B consistently achieve higher accuracy across all grades. Smaller models (1.5B-3B) perform poorly, particularly on lower grade levels (Grades 1-5), where their accuracy ranges between 50-60\%. Their performance improves to 70-80\% on higher grades, indicating difficulty in adapting to simpler, age-appropriate content. Mid-sized models like Qwen2.5-7B and LLaMA3-8B perform significantly better than smaller models and are often comparable to large models. Notably, the Qwen series consistently outperforms other models of the same size, with Qwen2.5-14B performing on par with the much larger Mistral-Small-24B. Although MCQs are relatively constrained and should be easier to answer, many models, especially smaller ones, still underperform. This highlights the diversity and challenge of our dataset, demonstrating gaps in current LLMs’ ability to handle grade-specific educational content.
\begin{table}[!t]
    \centering
    \scalebox{0.70}{
    {\renewcommand{\arraystretch}{1.3}}
    \begin{tabular}{l c c c c}
    \toprule
    \textbf{Model} & \textbf{Grade 1-2} & \textbf{Grade 3-5} & \textbf{Grade 6-8} & \textbf{Grade 9-12} \\
    \midrule
    qwen2.5-1.5B    & 54.2 & 55.5 & 58.9 & 67.3 \\
    SmolLM-1.7B     & 37.4 & 38.3 & 65.1 & 73.0 \\
    gemma-2b-it     & 41.2 & 36.5 & 50.1 & 60.9 \\
    qwen2.5-3B      & 69.3 & 70.4 & 70.0 & 78.9 \\
    llama3.2-3B     & 62.7 & 51.6 & 68.8 & 75.6 \\
    qwen2.5-7B      & \textbf{92.0} & 83.5 & 74.2 & 83.8 \\
    llama3-8B       & 64.2 & 68.3 & 75.1 & 80.4 \\
    qwen2.5-14B     & 83.1 & 84.4 & 78.6 & 85.9 \\
    mistral24B      & 86.2 & \textbf{85.8} & \textbf{80.5} & \textbf{86.3} \\
    \bottomrule
    \end{tabular}
    }
    \caption{Model-wise accuracy (\%) on MCQ questions across grade levels.}
    \label{tab:mcq_accuracy_compact}
\end{table}

\paragraph{Results on open-ended questions set.} 
As shown in Table~\ref{tab:judge_table}, all models, regardless of size, struggle to generate grade-appropriate responses for lower grades (Grades 1-5) compared to higher grades. Larger models like Qwen2.5-14B and Mistral-Small-24B consistently outperform others across all levels but still exhibit weaker performance on early-grade content. Mid-sized models, such as Qwen2.5-7B and LLaMA3-8B, perform slightly below the large models and follow a similar trend of reduced effectiveness in lower grades. Smaller models (1.5B-3B) perform noticeably worse than both mid-sized and large models across all grade levels. However, their performance gradually improves as the grade level increases, indicating better alignment with higher-grade content despite overall lower effectiveness.

\paragraph{Analysis.} The results reveal a clear gap: LLMs are generally better aligned with content for higher-grade students and struggle to adapt to early-grade needs. \textsc{EduAdapt} addresses this by providing the first benchmark that spans the full K-12 range, enabling systematic evaluation of grade-level adaptability.  Beyond LLM-based judgments, we also tested standard automated metrics, BLEU \cite{10.3115/1073083.1073135}, ROUGE (ROUGE-1, ROUGE-2, ROUGE-L) \cite{lin-2004-rouge}, and BERTScore \cite{zhang2020bertscoreevaluatingtextgeneration} - commonly used in text generation to measure overlap between generated and reference answers (Table~\ref{tab:trad_metrices} in Appendix~\ref{sec:AppendixB}). However, these metrics proved unreliable for educational QA. For example, BERTScore often exceeded 90\% even when answers were semantically wrong or developmentally inappropriate. Similarly, BLEU and ROUGE produced flat distributions that failed to capture differences in correctness or grade alignment.  

This contrast highlights the limits of traditional metrics: they capture surface-level similarity but overlook deeper qualities such as appropriateness, developmental fit, subject relevance, and factual accuracy. In comparison, accuracy for MCQs and LLM-as-a-judge scoring for open-ended responses provide more meaningful, pedagogically grounded assessments.

\begin{table}[!t]
\setlength{\tabcolsep}{3pt}
\renewcommand{\arraystretch}{1.4}
\scriptsize
\centering
\resizebox{\columnwidth}{!}{
\begin{tabular}{l c c c c }
\toprule
\textbf{Models} & \textbf{Gpt-4o} & \textbf{Qwen2.5-72B}  & \textbf{Llama3.3-70B} & \textbf{Average}\\
\midrule
\multicolumn{4}{c}{\textbf{Grade 1-2}} \\
qwen2.5-1.5B    & 4.36 & 5.42 & 5.32 & 5.03 \\
SmolLM-1.7B     & 4.25 & 5.36 & 5.33 & 4.98\\
gemma-2b-it     & 3.35 & 4.34 & 4.35 & 4.01\\
qwen2.5-3B      & 4.40 & 5.96 & 5.52 & 5.29\\
llama3.2-3B     & 4.36 & 5.37 & 5.48 & 5.07\\
qwen2.5-7B      & 4.99 & 6.34 & 6.09 & 5.80\\
llama3-8B       & 4.77 & 5.75 & 5.69 & 5.40\\
qwen2.5-14B     & 5.41 & \textbf{6.57} & 6.55 & 6.17\\
mistral24B      & \textbf{5.53} & 6.55 & \textbf{6.59} & \textbf{6.22}\\
\midrule
\midrule
\multicolumn{4}{c}{\textbf{Grade 3-5}} \\
qwen2.5-1.5B  & 4.84 & 5.95 & 5.84 & 5.54 \\
SmolLM-1.7B &  5.27   &  6.32       &   6.34 & 5.97     \\
gemma-2b-it    &  4.18   & 5.14 &   5.13  & 4.81        \\
qwen2.5-3B   &    5.30    & 6.79   &  6.49 & 6.19    \\
llama3.2-3B    & 5.06     & 6.22   & 6.39 & 5.89       \\
qwen2.5-7B     & 5.84     & 7.01   & 6.93  & 6.59    \\
llama3-8B      & 5.70     & 6.77   & 6.83  &  6.43        \\
qwen2.5-14B   & 6.31     & 7.22  & 7.20  & 6.91      \\
mistral24B     & \textbf{6.46 }    & \textbf{7.44 }    & \textbf{7.37} & \textbf{7.09} \\
\midrule
\midrule
\multicolumn{4}{c}{\textbf{Grade 6-8}} \\
qwen2.5-1.5B    &   4.89    &   5.89    &  5.74  & 5.50  \\
SmolLM-1.7B &  5.47   &  6.51       &   6.32 & 6.10     \\
gemma-2b-it    &  3.96   & 5.01 &   5.41  & 4.79        \\
qwen2.5-3B   &    5.67    & 7.00   &  6.92 & 6.53    \\
llama3.2-3B    & 5.96     & 6.88   & 7.17 & 6.67       \\
qwen2.5-7B     & 6.14     & 7.42   & 7.59 & 7.05     \\
llama3-8B      & 6.60     & 7.51   & 7.49 & 7.20          \\
qwen2.5-14B   & 6.51     & 7.86  & 7.98  & 7.45      \\
mistral24B     & \textbf{6.96 }    & \textbf{8.05}     & \textbf{8.09} & \textbf{7.70 }       \\
\midrule
\midrule
\multicolumn{4}{c}{\textbf{Grade 9-12}} \\
qwen2.5-1.5B    &   5.84    &   6.73    &  6.78  & 6.45  \\
SmolLM-1.7B &  5.96   &  7.30       &   7.42  & 6.89    \\
gemma-2b-it    &  4.88   & 6.09 &   5.72 & 5.56         \\
qwen2.5-3B   &    6.91    & 7.89   &  7.97  & 7.59   \\
llama3.2-3B    & 6.76     & 7.56   & 7.32 & 7.21       \\
qwen2.5-7B     & 7.54     & 8.38   & 8.56  & 8.16    \\
llama3-8B      & 6.74     & 8.04   & 8.12 & 7.63         \\
qwen2.5-14B   & 7.68     & 8.59  & 8.27 & 8.18       \\
mistral24B     & \textbf{7.94}     & \textbf{8.63}     & \textbf{8.58 }     & \textbf{8.38}   \\
     
\bottomrule
\end{tabular}}
\caption{Model-wise average scores on a scale of 1-10 for open-ended QA tasks across grade levels, as evaluated by three LLM judges: \textbf{GPT-4o}, \textbf{Qwen2.5-72B}, and \textbf{Llama3.3-70B}. The scores reflect the alignment of each model’s responses with grade-specific expectations. Bold values mark the highest score per grade level, highlighting the model judged most aligned with the target grade.}
\label{tab:judge_table}
\end{table}

\paragraph{Error Analysis of Model-Generated Responses.} To understand why the models generally struggled on our grade-appropriate test sets, we carried out a qualitative error analysis of their responses. This goes beyond numerical scores to examine the actual answers, highlighting the recurring issues, error patterns, and grade-specific challenges that explain their weaker performance. We analyzed \textbf{60} randomly sampled model responses across all subjects: \textbf{20} from Gemma-2B (5 per grade level), \textbf{20} from LLaMA3-8B, and \textbf{20} from Mistral-Small-24B. Key observations are summarized below, with one exemplar per grade shown in Listings~\ref{listing:gemma}, \ref{listing:llama}, and \ref{listing:mistral} in Appendix~\ref{sec:AppendixF}. The overall distribution of error types across models and grade levels is summarized in Table~\ref{tab:error_patterns}.

\textbf{Gemma-2B.} At Grades 1–2, performance was inconsistent, with many recall questions leading to hallucinations and partial misalignments. In Grades 3-5, clarity was maintained, but correctness and focus drift were the main issues. At Grades 6-8, answers were fluent but leaned on fabricated detail, with elaboration replacing precision. By Grades 9-12, responses were clear in form but factually unreliable, with verbosity rising yet accuracy not improving. Most frequent deficits: \emph{hallucination, correctness, and conceptual alignment}, with strengths in clarity and fluency.

\textbf{LLaMA3-8B.} At Grades 1-2, responses were lively and age-appropriate but often imprecise, with omissions and imaginative elaborations causing correctness and hallucination issues. In Grades 3-5, child-friendly phrasing persisted, but exaggerated details heightened correctness, hallucination, and focus drift problems. At Grades 6-8, consistent factual errors undermined precision despite a readable style. By Grades 9-12, outputs were clear and structured yet dominated by fabricated explanations, showing severe correctness and alignment issues. Most frequent deficits: \emph{correctness, hallucination, and focus drift/conceptual alignment}, while strengths lie in clarity and accessibility.

\textbf{Mistral-Small-24B.} At Grades 1-2 and 3-5, responses were phrased in simple, age-appropriate language but suffered from weak factual reliability, with frequent hallucinations and focus drift overshadowing the concise gold points. In Grades 6-8, correctness was more acceptable, yet over-explaining and added narrative detail reduced precision and grade alignment. At Grades 9-12, answers were clear, structured, and scientifically phrased, though factual reliability and alignment remained inconsistent due to hallucination and partial omissions. Most frequent deficits: \emph{correctness, conceptual alignment, and completeness}, while strengths lay in readability, clarity, and appropriate phrasing.

\begin{table}[!t]
\setlength{\tabcolsep}{3pt}
\renewcommand{\arraystretch}{1.4}
\scriptsize
\centering
\resizebox{\columnwidth}{!}{
\begin{tabular}{l l c c c c c}
\toprule
\textbf{Model} & \textbf{Grade} & \textbf{Hal.} & \textbf{FD} & \textbf{Ver.} & \textbf{CO} & \textbf{CAI} \\
\midrule
\multirow{4}{*}{Gemma-2B}
& 1--2  & \rcir & \rcir & \ocir & \rcir & \rcir \\
& 3--5  & \rcir & \rcir & \ocir & \rcir & \rcir \\
& 6--8  & \rcir & \rcir & \ocir & \rcir & \rcir \\
& 9--12 & \rcir & \ycir & \ocir & \rcir & \rcir \\
\midrule
\multirow{4}{*}{LLaMA3-8B}
& 1--2  & \ocir & \ocir & \ycir & \ocir & \ocir \\
& 3--5  & \ocir & \ycir & \ocir & \ocir & \ocir \\
& 6--8  & \rcir & \ocir & \ocir & \rcir & \ocir \\
& 9--12 & \rcir & \ycir & \ocir & \rcir & \rcir \\
\midrule
\multirow{4}{*}{Mistral-Small-24B}
& 1--2  & \ocir & \ocir & \ycir & \rcir & \ocir \\
& 3--5  & \ocir & \ocir & \ocir & \rcir & \ocir \\
& 6--8  & \gcir & \ycir & \ocir & \ycir & \ocir \\
& 9--12 & \ocir & \ycir & \ycir & \ocir & \ycir \\
\bottomrule
\end{tabular}}
\caption{Qualitative error analysis on a sample set of model-generated responses across grade levels.
Columns: Hallucination (Hal.), Focus Drift (FD), Verbosity (Ver.), Correctness(CO),
Conceptual Alignment Issues (CAI). \textbf{Legend:} \gcir~No Issues , \ycir~Minor Issues, \ocir~Moderate Issues, \rcir~Serious Issues, \xmark~Critical Issues.}
\label{tab:error_patterns}
\end{table}

\section{Literature Review}

This section reviews benchmark datasets used to evaluate LLMs on educational tasks, with a focus on math-centric, interdisciplinary, and multilingual evaluations. These resources are essential for measuring model performance and guiding the design of AI systems that support diverse learning needs.  

\paragraph{Math-Centric Benchmarks.} In mathematics, several benchmarks assess reasoning across grade levels. GSM8K \cite{cobbe2021trainingverifierssolvemath} provides 8.5K grade school problems, while the MATH dataset \cite{hendrycks2021measuringmathematicalproblemsolving} extends coverage to high school competition-style questions with detailed solutions. Larger collections like Dolphin18K \cite{huang-etal-2016-well}, Math23K \cite{wang-etal-2017-deep}, and MathQA \cite{amini2019mathqainterpretablemathword} broaden scope with real-world or multi-choice word problems, while DRAW-1K \cite{upadhyay2017annotatingderivationsnewevaluation} emphasizes the evaluation of derivations. Together, these highlight both the opportunities and challenges of scaling math reasoning evaluation.  

\paragraph{Interdisciplinary and Domain-Specific Benchmarks.} Beyond math, specialized and interdisciplinary benchmarks capture domain-specific reasoning. MedMCQA \cite{pal2022medmcqalargescalemultisubject} contains 194K medical exam questions, while TheoremQA \cite{chen-etal-2023-theoremqa} tests application of scientific theorems across multiple fields. ARC \cite{clark2018thinksolvedquestionanswering} evaluates science reasoning for Grades 3-9, distinguishing between retrieval and deeper inference. In programming education, datasets such as Defects4J \cite{10.1145/2610384.2628055}, ManyBugs, and IntroClass \cite{7153570} support research on automated program repair, while CodeReviewer \cite{li2022automatingcodereviewactivities} and follow-up work \cite{guo2023exploringpotentialchatgptautomated} focus on code review. Other resources include SciQ \cite{welbl2017crowdsourcingmultiplechoicescience} with 13.7K science QA pairs and FairytaleQA \cite{xu2022fantasticquestionsthemfairytaleqa} with 10K comprehension pairs for children’s stories.  

\paragraph{Multilingual and Global Benchmarks.} Multilingual and global benchmarks further broaden evaluation. C-EVAL \cite{huang2023cevalmultilevelmultidisciplinechinese} and GAOKAO-Bench \cite{zhang2024evaluatingperformancelargelanguage} assess Chinese educational content, while AGIEval \cite{zhong2023agievalhumancentricbenchmarkevaluating} uses real-world exam questions such as the SAT and LSAT. MMLU \cite{hendrycks2021measuringmassivemultitasklanguage} and CMMLU \cite{li2024cmmlumeasuringmassivemultitask} cover dozens of academic subjects in English and Chinese, exposing weaknesses in multi-step reasoning and negation. More recent work such as MATH-Vision \cite{wang2024measuringmultimodalmathematicalreasoning} incorporates visual reasoning into math tasks.  

Despite these advances, existing benchmarks show that even state-of-the-art models struggle with STEM subjects, deeper reasoning, and adaptation across grade levels. Closing this gap requires benchmarks that capture both subject-specific reasoning and developmental appropriateness to better support diverse learners.

\section{Conclusion and Future Work}

This work introduces the first comprehensive synthetic benchmark for evaluating educational QA across all K-12 grade levels. High-quality QA pairs were validated through a combination of LLM-based and expert human review and used to assess a range of language models. Using accuracy for MCQs and LLM-as-a-Judge scoring for open-ended responses, we evaluated how well current open-source LLMs align with the linguistic and cognitive needs of students at different stages. Results show a clear performance gap: models struggle significantly with lower-grade content (Grades 1-5), achieving only 60-70\% on open-ended questions, compared to up to 85\% for higher grades. Smaller models, in particular, showed poor performance across both MCQs and open-ended qa. These findings underscore the need for grade-aware training, prompting, and fine-tuning strategies tailored to younger learners.

Looking ahead, several directions can further improve educational language models. Expanding subject coverage will enable broader curriculum alignment, while incorporating multimodal QA (e.g., image or diagram-based) will better reflect real-world assessments. Supporting multilingual QA will increase accessibility for non-English-speaking students. Finally, addressing lower-grade performance through data augmentation, curriculum-aligned pretraining, and targeted fine-tuning is critical. Together, these efforts aim to build more reliable pedagogically grounded educational AI systems.

\section*{Limitations}

Despite the contributions of this work, it is important to acknowledge several limitations that shape the scope of the findings and highlight directions for future improvements. First, the dataset shows an imbalance in grade-level distribution, with fewer qa pairs for lower grades (Grades 1--5) compared to upper grades (Grades 6-12). Future work should aim to create more balanced datasets to enable more comprehensive assessments across all grade levels. Second, the dataset is based on a K-12 curriculum framework, which may limit its generalizability to other educational systems. As curricular standards and cognitive expectations vary globally, adapting and extending the dataset for international contexts is essential for broader applicability. Finally, while subjects such as ecology and geography draw examples from diverse regions, the benchmark does not account for regional variations in curricular relevance. A question that is simple for students in one region may be unfamiliar or challenging in another. This work, therefore, represents an initial step toward designing benchmarks that measure LLM adaptability to grade-level responses in a region-agnostic manner, with future refinements needed for culturally and regionally grounded evaluation. These limitations point to key areas for refinement, including more balanced data generation, enhanced cross-curricular and cross-regional alignment, and deeper integration of human judgment to support more accurate and inclusive educational evaluations.

\section*{Ethical Statement}

Ethical responsibility was a core principle throughout this study. From data generation using Wikipedia articles to evaluating educational QA pairs, all stages were designed to ensure transparency, fairness, and minimal bias. Using publicly available sources like Wikipedia promoted reproducibility and avoided risks associated with private or sensitive data, aligning with the broader goals of openness and accountability in educational AI research.

Recognizing potential biases in LLMs due to pretraining data, we employed a diverse set of models varying in size and architecture to reduce reliance on any single system. For evaluation, we used an LLM-as-a-Judge framework with three independent models, supplemented by manual review to ensure reliability and consistency. No personal or identifiable student data was used. All generated content and evaluations were conducted solely for academic research. This study aims to contribute responsibly to the development of educational AI systems, emphasizing fairness, transparency, and trust.

\section*{Acknowledgments}\label{sec:acknow}
Muhammad Abdul-Mageed acknowledges support from Canada Research Chairs (CRC), the Natural Sciences and Engineering Research Council of Canada (NSERC; RGPIN-2018-04267), the Social Sciences and Humanities Research Council of Canada (SSHRC; 895-2020-1004; 895-2021-1008), Canadian Foundation for Innovation (CFI; 37771), Digital Research Alliance of Canada,\footnote{\href{https://alliancecan.ca}{https://alliancecan.ca}} and UBC Advanced Research Computing-Sockeye.\footnote{\href{https://arc.ubc.ca/ubc-arc-sockeye}{https://arc.ubc.ca/ubc-arc-sockeye}}
\bibliography{custom}
\appendix
\newpage
\onecolumn

\section{Prompts}
\subsection{Grade-Level Text Classification Prompt}
\label{subsec:appendixA.1}

\begin{listing*}[!h]
\small
    \begin{minted}[frame=lines,
framesep=0.7mm,
baselinestretch=0.9,
fontsize=\footnotesize,
mathescape, 
breaklines]{bash}
"You are an expert educational content reviewer.  Your task is to determine which school grade level the {text} is most appropriate for {subject}, based on the concepts, vocabulary, and developmental stage required to understand it.  

The possible grade levels are:  
- Grade 1-2 (very simple, basic concepts, everyday vocabulary, short sentences)  
- Grade 3-5 (slightly more advanced concepts, simple explanations of science/social studies, common academic words)  
- Grade 6-8 (moderately complex concepts, multi-step reasoning, domain-specific terms start appearing)  
- Grade 9-12 (abstract ideas, higher-order reasoning, subject-specific terminology, complex sentence structures)  

Instructions:  
1. Analyze the text for difficulty, conceptual depth, and vocabulary.  
2. Choose only one grade level from the list above."
\end{minted}
    \caption{Prompt for grade-level classification of text passages}
    \label{listing:classify}
\end{listing*}

\subsection{Grade-Level QA Generation Prompts}
\label{subsec:appendixA.2}

\begin{listing*}[!h]
\small
    \begin{minted}[frame=lines,
framesep=0.7mm,
baselinestretch=0.9,
fontsize=\footnotesize,
mathescape, 
breaklines]{bash}
"You are an AI assistant specializing in creating educational content for young learners. Your task is to generate two simple, question-answer (QA) pairs (one mcq type and one qa type) based on the given text, suitable for Grade 1 and 2 students.

Instructions:
    - Use simple, short sentences with easy vocabulary appropriate for 6-8 year-old children.
    - Ask about observable things (what something looks like, where it lives, etc.).
    - Avoid reasoning or multi-step thinking
    - Keep the tone friendly, fun, and age-appropriate."
\end{minted}
    \caption{Prompt for grade 1 and 2 question-answer generation}
    \label{listing:grade1_2}
\end{listing*}

\begin{listing*}[!h]
\small
    \begin{minted}[frame=lines,
framesep=0.7mm,
baselinestretch=0.9,
fontsize=\footnotesize,
mathescape, 
breaklines]{bash}
"You are an AI assistant specializing in creating educational content for students in Grades 3 to 5. Your task is to generate two question-answer (QA) pairs (one mcq type and one qa type) based on the given text.
Instructions:
    - Use clear language suitable for ages 8-11
    - Use clear and concise language. Avoid overly complex words, but encourage age-appropriate critical thinking and explanation.
    - Focus on helping students understand important facts and cause-and-effect relationships.
    - Encourage observational or factual reasoning, not abstract modeling.
    - Keep the tone engaging, educational, and appropriate for upper elementary school learners."
\end{minted}
    \caption{Prompt for grade 3 to 5 question-answer generation}
    \label{listing:grade3_5}
\end{listing*}

\begin{listing*}[!h]
    \begin{minted}[frame=lines,
framesep=0.7mm,
baselinestretch=0.9,
fontsize=\footnotesize,
mathescape, 
breaklines]{bash}
"You are a AI assistant specializing in creating a Question-Answer (QA) pair for middle school students (Grades 6 to 8) based on the provided text. Your task is to generate a {qa_or_mcq} based on the given text.
Instructions:
    - Use vocabulary and complexity suitable for students aged 12-14.
    - Ask questions that require students to interpret information, reason through cause-and-effect, apply models, or predict outcomes.
    - Focus on scientific relationships, system interactions, and basic modeling of processes or phenomena.
    - Simplify complex or abstract ideas into familiar contexts that students can reason about.
    - Maintain an educational tone that encourages scientific thinking and exploration."
\end{minted}
    \caption{Prompt for grade 6 to 8 question-answer generation}
    \label{listing:grade6_8}
\end{listing*}

\begin{listing*}[!h]
    \begin{minted}[frame=lines,
framesep=0.7mm,
baselinestretch=1,
fontsize=\footnotesize,
mathescape, 
breaklines]{bash}
"You are a AI assistant specializing in creating a Question-Answer (QA) pair for high school students (Grades 9 to 12) based on the provided text. Your task is to generate a {qa_or_mcq} based on the given text.
Instructions:
    - Use academically precise language appropriate for students aged 14-18 preparing for advanced or college-level studies.
    - Focus on modeling, applying laws, analyzing systems, and using quantitative or qualitative relationships.
    - Ask questions that require students to analyze, model, predict, calculate, or critically evaluate scenarios.
    - Ensure the question and answer are fully self-contained and understandable without needing to reference the original text.
    - Maintain an academic, analytical tone suited for high school science learners."
\end{minted}
    \caption{Prompt for grade 9 to 12 question-answer generation}
    \label{listing:grade9_12}
\end{listing*}

\subsection{Self-Reflection and Human Evaluation Prompts}
\label{subsec:appendixA.3}

\begin{listing}[!h]
    \begin{minted}[frame=lines,
framesep=0.7mm,
baselinestretch=1,
fontsize=\footnotesize,
mathescape, 
breaklines]{bash}
"Your role is to evaluate each question-answer pair for Grade {grade_level} students in the subject of {subject}.

You must rate each qa pair on the following criteria, each on a scale of 1-10:

Evaluation Criteria:

    1. language-appropriateness: Is the language simple, short, and easy for 6-8 year-old children to understand?
    2. grade-alignment: Does the question reflect what students at this age typically observe or experience.
    3. relevance: Is the question-answer pair based on observable actions or phenomena, and understandable on its own without needing to refer back to any source?
    4. clarity: Is the question phrased clearly, with an unambiguous answer?
    5. subject-fit ({subject}): Does the question relate to age-appropriate scientific concepts in this subject, without factual inaccuracies or misconceptions?"
\end{minted}
    \caption{Prompt for evaluating the quality of grade 1-2 QA pairs through self-reflection}
    \label{listing:grade_1_2_self-reflection_prompt}
\end{listing}

\begin{listing}[!h]
    \begin{minted}[frame=lines,
framesep=0.7mm,
baselinestretch=1.1,
fontsize=\footnotesize,
mathescape, 
breaklines]{bash}
"Your role is to evaluate each question-answer pair for Grade {grade_level} students in the subject of {subject}.

You must rate each qa pair on the following criteria, each on a scale of 1-10:

Evaluation Criteria:

    1. language-appropriateness: Is the language clear, age-appropriate (for 8–11-year-old students), avoiding overly complex vocabulary but encouraging basic reasoning?
    2. grade-alignment: Does the question match the cognitive and curriculum expectations for Grades 3–5, focusing on understanding facts, cause-and-effect, or simple scientific reasoning?
    3. relevance: Is the QA pair directly related to observable phenomena, simple explanations, or important scientific facts appropriate to the grade level?
    4. clarity: Is the question clearly phrased, guiding students to provide or recognize a straightforward explanation or prediction?
    5. subject-fit ({subject}): Does the content accurately reflect important concepts from the subject suitable for upper elementary learners?"
\end{minted}
    \caption{Prompt for evaluating the quality of grade 3-5 QA pairs through self-reflection}
    \label{listing:grade_3_5_self-reflection_prompt}
\end{listing}

\newpage

\begin{listing}[!h]
    \begin{minted}[frame=lines,
framesep=0.7mm,
baselinestretch=0.9,
fontsize=\footnotesize,
mathescape, 
breaklines]{bash}
"Your role is to evaluate each question-answer pair for Grade {grade_level} students in the subject of {subject}.

You must rate each qa pair on the following criteria, each on a scale of 1-10:


Evaluation Criteria:

    1. language-appropriateness: Is the language clear, precise, and appropriate for 12-14 year-old students, supporting intermediate scientific reasoning?
    2. grade-alignment: Does the question match the cognitive expectations for middle school learners, involving interpretation, cause-and-effect analysis, simple system modeling, or predictions?
    3. relevance: Is the QA pair rooted in scientific phenomena, relationships, or system-level interactions appropriate to the grade level?
    4. clarity: Is the question phrased clearly, guiding students to reason, analyze, or predict in a focused and understandable way?
    5. subject-fit ({subject}): Does the content reflect accurate and important scientific concepts appropriate for middle school science in this subject?
    "
\end{minted}
    \caption{Prompt for evaluating the quality of grade 6-8 QA pairs through self-reflection}
    \label{listing:grade_6_8_self-reflection_prompt}
\end{listing}

\begin{listing}[!h]
    \begin{minted}[frame=lines,
framesep=0.7mm,
baselinestretch=0.9,
fontsize=\footnotesize,
mathescape, 
breaklines]{bash}
"Your role is to evaluate each question-answer pair for Grade {grade_level} students in the subject of {subject}.

You must rate each qa pair on the following criteria, each on a scale of 1-10:

Evaluation Criteria:

    1. language-appropriateness: Is the language academically precise and appropriate for students aged 14-18 preparing for advanced science studies?
    2. grade-alignment: Does the question meet the cognitive expectations for high school learners, requiring multi-step reasoning, quantitative analysis, modeling, or critical evaluation?
    3. relevance: Is the QA pair grounded in substantial scientific concepts, systems modeling, or data-driven explanations appropriate for high school science?
    4. clarity: Is the question phrased clearly and at a cognitive depth suitable for high school students?
    5. subject-fit ({subject}): Does the content align with advanced high school curriculum topics within the subject, and maintain scientific accuracy?
"
\end{minted}
    \caption{Prompt for evaluating the quality of grade 9-12 QA pairs through self-reflection}
    \label{listing:grade_9_12_self-reflection_prompt}
\end{listing}

\subsection{Prompt for LLM-as-a-Judge Evaluation}
\label{subsec:appendixA.4}
\begin{listing}[!h]
    \begin{minted}[frame=lines,
framesep=0.7mm,
baselinestretch=0.9,
fontsize=\footnotesize,
mathescape, 
breaklines]{bash}
"""Your role is to evaluate the model’s response for a student of Grade {{ grade_level }} by comparing it to the gold answer.

Evaluation Criteria:
  Evaluate the model's response in relation to the gold answer, based on the following criteria:

  - Vocabulary Alignment: Does the model use vocabulary that closely matches the complexity, accessibility, and tone of the gold answer, assuming the gold answer is already grade-appropriate?
  - Conceptual Alignment: Does the model's response reflect a similar level of cognitive and conceptual depth as the gold answer?
  - Scientific Language Alignment: Does the model use scientific or technical terms in a way that aligns with the gold answer in terms of complexity and usage?
  - Correctness: Is the model's answer factually accurate and consistent with the gold answer?
  - Clarity: Is the model's response as clear, coherent, and well-structured as the gold answer?
  - Completeness: Does the model's answer cover the same key ideas, explanations, or observations as the gold answer?

  Assign a rating from 1 to 10 to each criteria based on how well the model's answer aligns with the gold answer across each criterion:
"""
\end{minted}
    \caption{Prompt for LLM-as-a-Judge evaluation}
    \label{listing:judge_prompt}
\end{listing}

\subsection{Prompt for Answer Generation from LLMs}
\label{subsec:appendixA.5}

\begin{listing}[!h]
    \begin{minted}[frame=lines,
framesep=0.7mm,
baselinestretch=1.2,
fontsize=\footnotesize,
mathescape, 
breaklines]{bash}
"You are an experienced educator answering questions for students in {grade_level}. Please give a clear and developmentally appropriate answer to the question below."
\end{minted}
    \caption{Prompt for evaluating different LLMs on testset}
    \label{listing:mode_evaluation_prompt}
\end{listing}

\section{Tables}
\label{sec:AppendixB}

\setlength{\tabcolsep}{3pt}
\begin{table}[!h]
\renewcommand{\arraystretch}{1.4}
\scriptsize
\centering
\resizebox{\columnwidth}{!}{
\begin{tabular}{ l c c  c  c  c  c c c  c c} 
\toprule
    {} & \textbf{Grade Levels} & \textbf{Biology} & \textbf{Physics} & \textbf{Chemistry} & \textbf{Computer Science} & \textbf{Ecology} & \textbf{Geography} & \textbf{Geology} & \textbf{Medicine} & \textbf{Metrology} \\
    \midrule
    & 1 and 2 & 277 & 48 & 75 & 92 & 474 & 732 & 159 & 85 & 55 \\
    \multirow{-9}{*}{\rotatebox[origin=c]{90}{\parbox[c]{4cm}{\texttt{\textbf{QA Count}~~~}}}} &   3 to 5  & 602 & 59 & 158 & 263 & 906 & 1252 & 483 & 236 & 127 \\ 
    & 6 to 8 & 206 & 55 & 90 & 185 & 440 & 272 & 180 & 165 & 132 \\
    & 9 to 12 & 844 & 1485 & 1624 & 1238 & 696 & 412 & 749 & 973 & 996 \\
    \midrule
    & 1 and 2 & 114 & 26 & 62 & 28 & 137 & 156 & 19 & 36 & 24  \\
    \multirow{-10}{*}{\rotatebox[origin=c]{90}{\parbox[c]{4.5cm}{\texttt{\textbf{MCQ Count}~~~}}}} &   3 to 5  & 217 & 28 & 49 & 68 & 419 & 543 & 145 & 44 & 36 \\
    & 6 to 8 & 245 & 57 & 88 & 184 & 572 & 284 & 250 & 112 & 73 \\
    & 9 to 12 & 675 & 1183 & 1453 & 1190 & 648 & 350 & 794 & 751 & 1111 \\
    
\midrule 
    \rotatebox[origin=c]{90}{\parbox[c]{1cm}{\texttt{\textbf{Total}}}} &   {}  & \textbf{3180} & \textbf{2941} & \textbf{3599} & \textbf{3248} & \textbf{4292} & \textbf{4001} & \textbf{2779} & \textbf{2402} & \textbf{2554}  \\ 
\bottomrule

\end{tabular} }

\caption{\textbf{Training Set:} Subject-wise and grade-level distribution of question-answer pairs in the training split.}
\label{table:train_data}
\end{table}

\setlength{\tabcolsep}{3pt}
\begin{table}[!h]
\renewcommand{\arraystretch}{1.4}
\scriptsize
\centering
\resizebox{\columnwidth}{!}{
\begin{tabular}{ l c c  c  c  c c c  c  c c } 
\toprule
    {} & \textbf{Grade Levels} & \textbf{Biology} & \textbf{Physics} & \textbf{Chemistry} & \textbf{Computer Science} & \textbf{Ecology}  & \textbf{Geography} & \textbf{Geology} & \textbf{Medicine} & \textbf{Metrology} \\
    \midrule
    & 1 and 2 & 92 & 16 & 25 & 30 & 158 & 244 & 44 & 31 & 15  \\
    \multirow{-9}{*}{\rotatebox[origin=c]{90}{\parbox[c]{4cm}{\texttt{\textbf{QA Count}~~~}}}} &   3 to 5  & 200 & 20 & 51 & 86 & 302 & 417 & 158 & 71 & 45 \\ 
    & 6 to 8 & 68 & 20 & 24 & 64 & 172 & 109 & 61 & 55 & 44 \\
    & 9 to 12 & 281 & 495 & 541 & 413 & 225 & 136 & 250 & 324 & 332 \\
    \midrule
    & 1 and 2 & 38 & 8 & 20 & 9 & 44 & 45 & 15 & 8 & 10 \\
    \multirow{-10}{*}{\rotatebox[origin=c]{90}{\parbox[c]{4.5cm}{\texttt{\textbf{MCQ Count}~~~}}}} &   3 to 5  & 72 & 8 & 18 & 23 & 152 & 187 & 51 & 21 & 8 \\ 
    & 6 to 8 & 81 & 16 & 34 & 58 & 164 & 76 & 82 & 41 & 23 \\
    & 9 to 12 & 225 & 395 & 484 & 397 & 227 & 119 & 265 & 250 & 370 \\
    
\midrule 
    \rotatebox[origin=c]{90}{\parbox[c]{1cm}{\texttt{\textbf{Total}}}} &   {}  & \textbf{1059} & \textbf{978} & \textbf{1197} & \textbf{1080} & \textbf{1444} & \textbf{1333} & \textbf{926} & \textbf{801} & \textbf{846} \\ 
\bottomrule
\end{tabular} }
\caption{\textbf{Development Set:} Subject-wise and grade-level distribution of question-answer pairs in the development split.}
\label{table:dev_data}
\end{table}

\newpage
\setlength{\tabcolsep}{3pt}
\begin{table}[!h]
\renewcommand{\arraystretch}{1.4}
\scriptsize
\centering
\resizebox{\columnwidth}{!}{
\begin{tabular}{ l c c  c  c  c  c c c   c c} 
\toprule
    {} & \textbf{Grade Levels} & \textbf{Biology} & \textbf{Physics} & \textbf{Chemistry} & \textbf{Computer Science} & \textbf{Ecology} & \textbf{Geography} & \textbf{Geology} & \textbf{Medicine} & \textbf{Metrology}\\
    \midrule
    & 1 and 2 & 93 & 16 & 25 & 32 & 158 & 244 & 53 & 30 & 22 \\
    \multirow{-14}{*}{\rotatebox[origin=c]{90}{\parbox[c]{6cm}{\texttt{\textbf{QA Count}~~~}}}} &   3 to 5  & 202 & 21 & 54 & 89 & 302 & 417 & 169 & 72 & 37\\ 
    & 6 to 8 & 70 & 14 & 30 & 53 & 147 & 82 & 58 & 55 & 45 \\
    & 9 to 12 & 282 & 495 & 541 & 413 & 222 & 132 & 250 & 324 & 332\\
    \midrule
    & 1 and 2 & 39 & 10 & 22 & 11 & 44 & 47 & 6 & 9 & 3 \\
    \multirow{-10}{*}{\rotatebox[origin=c]{90}{\parbox[c]{4.5cm}{\texttt{\textbf{MCQ Count}~~~}}}} &   3 to 5  & 74 & 7 & 15 & 20 & 165 & 183 & 40 & 20 & 16\\ 
    & 6 to 8 & 83 & 22 & 28 & 69 & 189 & 103 & 85 & 34 & 22 \\
    & 9 to 12 & 225 & 395 & 484 & 397 & 230 & 123 & 265 & 250 & 370 \\
    
\midrule 
    \rotatebox[origin=c]{90}{\parbox[c]{1cm}{\texttt{\textbf{Total}}}} &   {}  & \textbf{1068} & \textbf{980} & \textbf{1199} & \textbf{1084} & \textbf{1457} & \textbf{1331} & \textbf{926} & \textbf{794} & \textbf{847} \\ 
\bottomrule

\end{tabular} }
\caption{\textbf{Test Set:} Subject-wise and grade-level distribution of question-answer pairs in the test split.}
\label{table:test_data}

\end{table}

\setlength{\tabcolsep}{3pt}
\begin{table}[!h]
\renewcommand{\arraystretch}{1.4}
\scriptsize
    \centering
    \resizebox{\columnwidth}{!}{\
    \begin{tabular}{L{3cm} L{5cm} L{3.3cm} }
    \toprule
      \textbf{Models} & \textbf{Huggingface Identifiers} & \textbf{Usage} \\
    \midrule \midrule
    phi-4  & microsoft/phi-4   & QA Generation/Self-Reflection                     \\ \hline
      Qwen2.5-1.5B                 & Qwen/Qwen2.5-1.5B-Instruct             & Evaluation         \\ \hline
       SmolLM-1.7B                 & HuggingFaceTB/SmolLM-1.7B-Instruct            & Evaluation         \\ \hline
       Gemma-2b-it                & google/gemma-2b-it            & Evaluation         \\ \hline
       Qwen2.5-3B                 & Qwen/Qwen2.5-3B-Instruct            & Evaluation         \\ \hline
        Llama3.2-3B                & meta-llama/Llama-3.2-3B-Instruct            & Evaluation         \\ \hline
       Qwen2.5-7B                 & Qwen/Qwen2.5-7B-Instruct            & Evaluation         \\ \hline
       Llama3-8B                & meta-llama/Meta-Llama-3-8B-Instruct            & Evaluation         \\ \hline
       Qwen2.5-14B                 & Qwen/Qwen2.5-14B-Instruct            & Evaluation         \\ \hline
       Mistral24B  & mistralai/Mistral-Small-24B-Instruct-2501 & Evaluation \\ \hline
       GPT4o  & gpt-4o-2024-08-06 & LLM-as-a-Judge \\ \hline
       Qwen2.5-72B  & Qwen/Qwen2.5-72B-Instruct-GPTQ-Int8 & LLM-as-a-Judge \\ \hline
       Llama-3.3-70B & meta-llama/Llama-3.3-70B-Instruct & LLM-as-a-Judge \\ \hline
       
    \bottomrule
         & 
    \end{tabular}}
    \caption{Huggingface identifiers of our models and their usage point across the pipeline}
    \label{tab:model_table}
\end{table}

\begin{table}[!h]
    \centering
    \scalebox{0.9}{
    \begin{tabular}{l c c c c}
    \toprule
    \textbf{Criteria} & \textbf{Grade 1--2} & \textbf{Grade 3--5} & \textbf{Grade 6--8} & \textbf{Grade 9--12} \\
    \midrule
    Language Appropriateness & 7.24 & 7.49 & 7.91 & 8.77 \\
    Grade Alignment          & 7.17 & 7.62 & 7.64 & 8.57 \\
    Relevance                & 8.06 & 8.17 & 8.14 & 8.95 \\
    Clarity                  & 8.16 & 8.68 & 8.57 & 8.82 \\
    Subject Fit              & 7.83 & 8.45 & 8.61 & 9.16 \\
    \bottomrule
    \end{tabular}
    }
    \caption{Average human evaluation scores across grade levels for each criterion.}
    \label{tab:avg_across_human}
\end{table}


\begin{table}[!t]
\setlength{\tabcolsep}{3pt}
\renewcommand{\arraystretch}{1.3}
\scriptsize
\centering
\resizebox{\columnwidth}{!}{
\begin{tabular}{l c c c c c c c c}
\toprule
\textbf{Models} & \textbf{ROUGE-1} & \textbf{ROUGE-2}  & \textbf{ROUGE-L} & \textbf{BERTScore} & \textbf{BLEUScore} & \textbf{Gpt-4o} & \textbf{Qwen2.5-72B} & \textbf{Llama3.3-70B} \\
\midrule
\multicolumn{9}{c}{\textbf{Grade 1–2}} \\
qwen2.5-1.5B	& 0.224 & 0.104 & 0.215 & 0.975 & 0.062 & 0.436 & 0.542 & 0.532 \\
SmolLM-1.7B 	& 0.110 & 0.042 & 0.102 & 0.965 & 0.019 & 0.425 & 0.536 & 0.533\\
gemma-2b-it 	& 0.109 & 0.032 & 0.098 & 0.965 & 0.014 & 0.335 & 0.434 & 0.435\\
qwen2.5-3B  	& 0.164 & 0.067 & 0.152 & 0.970 & 0.033 & 0.440 & 0.596 & 0.552\\
llama3.2-3B 	& 0.178 & 0.070 & 0.162 & 0.969 & 0.037 & 0.436 & 0.537 & 0.548\\
qwen2.5-7B  	& 0.162 & 0.060 & 0.149 & 0.969 & 0.031 & 0.499 & 0.634 & 0.609\\
llama3-8B   	& 0.135 & 0.051 & 0.127 & 0.965 & 0.022 & 0.477 & 0.575 & 0.569\\
qwen2.5-14B 	& 0.157 & 0.069 & 0.144 & 0.970 & 0.024 & 0.541 & \textbf{0.657 }& 0.655\\
mistral24B  	& 0.132 & 0.044 & 0.121 & 0.966 & 0.0167 & \textbf{0.553} & 0.655 & \textbf{0.659}\\
\midrule
\midrule
\multicolumn{9}{c}{\textbf{Grade 3-5}} \\
qwen2.5-1.5B  & 0.263 & 0.155 & 0.0.233 & 0.966 & 0.100 & 0.484 & 0.595 & 0.584 \\
SmolLM-1.7B &  0.213   &  0.117  &  0.192 & 0.966 & 0.061 & 0.527 & 0.632 & 0.634\\    
gemma-2b-it    &  0.217   & 0.110 &   0.192  & 0.965 & 0.060   & 0.418 & 0.514 & 0.513   \\
qwen2.5-3B   &    0.107    & 0.048   &  0.093 & 0.956 & 0.022  & 0.530 & 0.679 & 0.649\\
llama3.2-3B    & 0.177     & 0.098   & 0.163 & 0.962  & 0.053  & 0.506 &0.622 & 0.639   \\
qwen2.5-7B     & 0.159     & 0.095   & 0.141  & 0.963 & 0.040 & 0.584 & 0.701 & 0.693  \\
llama3-8B      & 0.144     & 0.071   & 0.129  &  0.958& 0.030   & 0.570 & 0.677 & 0.683     \\
qwen2.5-14B   & 0.169     & 0.080  & 0.148  & 0.969  & 0.038  & 0.631 & 0.722 & 0.720  \\
mistral24B     & 0.180     & 0.092     & 0.157 & 0.962 & 0.044 & \textbf{0.646} & \textbf{0.744} & \textbf{0.737} \\
\midrule
\midrule
\multicolumn{9}{c}{\textbf{Grade 6-8}} \\
qwen2.5-1.5B    &   0.314    &   0.149    &  0.963  & 0.550  & 0.085 & 0.489 & 0.589 & 0.574\\
SmolLM-1.7B &  0.291   &  0.140       &   0.225 & 0.969 & 0.075  & 0.547 & 0.651 & 0.632   \\
gemma-2b-it    &  0.239   & 0.099 &   0.177  & 0.959  & 0.049   & 0.396 & 0.501 & 0.541   \\
qwen2.5-3B   &    0.173    & 0.064   &  0.127 & 0.950  & 0.029 & 0.567 & 0.700 & 0.692 \\
llama3.2-3B    & 0.191     & 0.084   & 0.146 & 0.960  & 0.038  & 0.596 & 0.688 & 0.717   \\
qwen2.5-7B     & 0.191     & 0.074   & 0.139 & 0.955 & 0.030  & 0.614 & 0.742 & 0.759   \\
llama3-8B      & 0.202     & 0.093   & 0.153 & 0.960 & 0.041  & 0.660 & 0.751 & 0.749       \\
qwen2.5-14B   & 0.244     & 0.105  & 0.181  & 0.960 & 0.048  & 0.651 & 0.786 & 0.798   \\
mistral24B     & 0.250     & 0.111     & 0.184 & 0.960 & 0.050   & \textbf{0.696} & \textbf{0.805} & \textbf{0.809}     \\
\midrule
\midrule
\multicolumn{9}{c}{\textbf{Grade 9-12}} \\
qwen2.5-1.5B    &   0.318    &   0.131    &  0.331  & 0.961  & 0.072 & 0.584 & 0.673 & 0.678\\
SmolLM-1.7B &  0.316   &  0.136       &   0.223  & 0. 961  & 0.069 & 0.596 & 0.730 & 0.742 \\
gemma-2b-it    &  0.262   & 0.099 &   0.177 & 0.960 & 0.043  & 0.488 & 0.609 & 0.572      \\
qwen2.5-3B   &    0.219    & 0.075   &  0.144  & 0.951 & 0.032 & 0.691 & 0.789 & 0.797  \\
llama3.2-3B    & 0.233     & 0.096   & 0.162 & 0.951  & 0.041 & 0.676 & 0.756 & 0.732    \\
qwen2.5-7B     & 0.232     & 0.096   & 0.154  & 0.951 & 0.039 & 0.754 & 0.838 & 0.856  \\
llama3-8B      & 0.253     & 0.108   & 0.174 & 0.951  & 0.045   & 0.674 & 0.804 & 0.812    \\
qwen2.5-14B   & 0.267     & 0.110  & 0.179 & 0.951  & 0.047  & 0.768 & 0.859 & 0.827   \\
mistral24B     & 0.281     & 0.114     & 0.951      & 0.838  & 0.047 & \textbf{0.794} & \textbf{0.863} & \textbf{0.858 }\\
     
\bottomrule
\end{tabular}}
\caption{Model-wise average scores for open-ended QA tasks across grade levels, as evaluated by three LLM judges,  \textbf{ROUGE-1}, \textbf{ROUGE-2}, and \textbf{ROUGE-L}. The scores reflect the alignment of each model’s responses with grade-specific expectations. Scores from the LLM judges are normalized to a 0-1 scale for comparability with standard metric values.}
\label{tab:trad_metrices}
\end{table}
\break

\newpage
\section{Grade-Level QA Pairs}
The sections below contain example qa pairs from our dataset for each field and grade level.
\label{sec:AppendixC}
\subsection{Biology}
\begin{listing}[!h]
\small
    \begin{minted}[frame=lines,
framesep=0.7mm,
baselinestretch=1.1,
fontsize=\footnotesize,
mathescape, 
breaklines]{bash}
"grade_level: grade 1 and 2
question: What kind of animals live in the Pigsties? 
answer: Pigs live in the Pigsties."
\end{minted}
    \caption{Grade 1 and 2 QA pair for biology}
    \label{listing:1_2_pair_bio}
\end{listing}

\begin{listing}[!h]
\small
    \begin{minted}[frame=lines,
framesep=0.7mm,
baselinestretch=1.1,
fontsize=\footnotesize,
mathescape, 
breaklines]{bash}
"grade_level: grade 3 to 5
question: What does it mean if a species is omnivorous?
answer: If a species is omnivorous, it means it eats both plants and animals. This allows the species to have a varied diet and adapt to different food sources available in its habitat."
\end{minted}
    \caption{Grade 3 to 5 QA pair for biology}
    \label{listing:3_5_pair_bio}
\end{listing}

\begin{listing}[!h]
\small
    \begin{minted}[frame=lines,
framesep=0.7mm,
baselinestretch=1.1,
fontsize=\footnotesize,
mathescape, 
breaklines]{bash}
"grade_level: grade 6 to 8
question: How do vampire bats locate blood vessels in their prey, and why is this adaptation important for their feeding habits?
answer: Vampire bats use heat sensors in their noses to detect blood vessels near the surface of the skin. This adaptation is important because it allows them to accurately find and target areas rich in blood, making their feeding process more efficient."
\end{minted}
    \caption{Grade 6 to 8 QA pair for biology}
    \label{listing:3_5_pair_bio}
\end{listing}

\begin{listing}[!h]
\small
    \begin{minted}[frame=lines,
framesep=0.7mm,
baselinestretch=1,
fontsize=\footnotesize,
mathescape, 
breaklines]{bash}
"grade_level: grade 9 to 12 
question: Explain why trypan blue is used as a vital stain in biosciences and how it helps differentiate between live and dead cells.
answer: Trypan blue is used as a vital stain in biosciences because it selectively colors dead tissues or cells blue, while live cells with intact cell membranes remain unstained. This is due to the selective permeability of cell membranes, which allows trypan blue to pass through and stain dead cells, but not live cells. This property makes it a useful tool for distinguishing between live and dead cells under a microscope, as dead cells appear blue while live cells do not take up the dye."
\end{minted}
    \caption{Grade 9 to 12 QA pair for biology}
    \label{listing:9_12_pair_bio}
\end{listing}

\subsection{Physics}

\begin{listing}[!h]
\small
    \begin{minted}[frame=lines,
framesep=0.7mm,
baselinestretch=1.1,
fontsize=\footnotesize,
mathescape, 
breaklines]{bash}
"grade_level: grade 1 and 2
question: What is a whirlpool? 
answer: A whirlpool is a swirling movement of water." 
\end{minted}
    \caption{Grade 1 and 2 QA pair for physics}
    \label{listing:1_2_pair_phy}
\end{listing}
\newpage
\begin{listing}[!h]
\small
    \begin{minted}[frame=lines,
framesep=0.7mm,
baselinestretch=1.1,
fontsize=\footnotesize,
mathescape, 
breaklines]{bash}
"grade_level: grade 3 to 5
question: How are sound waves in water detected by a receiver like the human ear or a hydrophone?
answer: Sound waves in water are detected by a receiver as changes in pressure. The receiver senses the alternating compressions and rarefactions of the water, which are changes in how tightly the water molecules are packed together."
\end{minted}
    \caption{Grade 3 to 5 QA pair for physics}
    \label{listing:3_5_pair_phy}
\end{listing}

\begin{listing}[!h]
\small
    \begin{minted}[frame=lines,
framesep=0.7mm,
baselinestretch=1.1,
fontsize=\footnotesize,
mathescape, 
breaklines]{bash}
"grade_level: grade 6 to 8
question: How does the pressure exerted by a glacier affect the melting point of ice at its base, and what is the result of this process?
answer: The pressure exerted by a glacier on its lower surface lowers the melting point of the ice, causing it to melt. This melting allows the glacier to move from a higher elevation to a lower elevation, and at lower elevations, the liquid water may flow from the base of the glacier when the air temperature is above the freezing point of water."
\end{minted}
    \caption{Grade 6 to 8 QA pair for physics}
    \label{listing:3_5_pair_phy}
\end{listing}

\begin{listing}[!h]
\small
    \begin{minted}[frame=lines,
framesep=0.7mm,
baselinestretch=1.1,
fontsize=\footnotesize,
mathescape, 
breaklines]{bash}
"grade_level: grade 9 to 12 
question: Explain how pressure piling can lead to a deflagration to detonation transition in connected vessels, and discuss the measures taken to prevent this in electrical equipment.
answer: Pressure piling occurs when a flame front propagates along a tube, compressing and heating the unburned gases ahead of it. This compression can significantly increase the pressure, ranging from twice to eight times the initial pressure. In systems where multiple vessels are connected by piping, this can lead to a deflagration to detonation transition, resulting in a very large explosion pressure. In electrical equipment in hazardous areas, this risk is mitigated by avoiding the use of conduits to connect classified equipment and by using barrier glands on cables entering enclosures. These measures ensure that compartments remain separate, preventing the transmission of explosions from one compartment to another."
\end{minted}
    \caption{Grade 9 to 12 QA pair for physics}
    \label{listing:9_12_pair_phy}
\end{listing}

\subsection{Chemistry}

\begin{listing}[!h]
\small
    \begin{minted}[frame=lines,
framesep=0.7mm,
baselinestretch=1.1,
fontsize=\footnotesize,
mathescape, 
breaklines]{bash}
"grade_level: grade 1 and 2
question: What does moisture mean? 
answer: Moisture means the presence of water, often in small amounts." 
\end{minted}
    \caption{Grade 1 and 2 QA pair for chemistry}
    \label{listing:1_2_pair_chem}
\end{listing}
\begin{listing}[!h]
\small
    \begin{minted}[frame=lines,
framesep=0.7mm,
baselinestretch=1.1,
fontsize=\footnotesize,
mathescape, 
breaklines]{bash}
"grade_level: grade 3 to 5
question: Why is it important to know if a substance is soluble in water?
answer: Knowing if a substance is soluble in water helps us understand how it can be used or handled. For example, if a substance dissolves in water, it can be mixed into drinks or used in cooking. It also helps scientists and engineers in creating solutions for cleaning, medicine, and other applications."
\end{minted}
    \caption{Grade 3 to 5 QA pair for chemistry}
    \label{listing:3_5_pair_chem}
\end{listing}
\newpage
\begin{listing}[!h]
\small
    \begin{minted}[frame=lines,
framesep=0.7mm,
baselinestretch=1.1,
fontsize=\footnotesize,
mathescape, 
breaklines]{bash}
"grade_level: grade 6 to 8
question: What is the difference between an accepted value and an experimental value in chemistry?
answer: An accepted value is a value of a substance's properties that is agreed upon by almost all scientists, while an experimental value is the value of a substance's properties that is determined in a specific laboratory setting."
\end{minted}
    \caption{Grade 6 to 8 QA pair for chemistry}
    \label{listing:6_8_pair_chem}
\end{listing}

\begin{listing}[!h]
\small
    \begin{minted}[frame=lines,
framesep=0.7mm,
baselinestretch=1.1,
fontsize=\footnotesize,
mathescape, 
breaklines]{bash}
"grade_level: grade 9 to 12 
question: Explain how acidosis affects the pH level of blood or body fluids, and why this change occurs.
answer: Acidosis increases the concentration of hydrogen ions in blood or body fluids. Since pH is the negative logarithm of hydrogen ion concentration, an increase in hydrogen ions results in a decrease in pH. This occurs because the pH scale is inversely related to hydrogen ion concentration; more hydrogen ions mean a lower pH, indicating increased acidity."
\end{minted}
    \caption{Grade 9 to 12 QA pair for chemistry}
    \label{listing:9_12_pair_chem}
\end{listing}

\subsection{Computer Science}

\begin{listing}[!h]
\small
    \begin{minted}[frame=lines,
framesep=0.7mm,
baselinestretch=1,
fontsize=\footnotesize,
mathescape, 
breaklines]{bash}
"grade_level: grade 1 and 2
question: What can you do with the Game Boy?
answer: "You can play games on the Game Boy." 
\end{minted}
    \caption{Grade 1 and 2 QA pair for computer science}
    \label{listing:1_2_pair_comp}
\end{listing}

\begin{listing}[!h]
\small
    \begin{minted}[frame=lines,
framesep=0.7mm,
baselinestretch=1,
fontsize=\footnotesize,
mathescape, 
breaklines]{bash}
"grade_level: grade 3 to 5
question: What is the main purpose of the Little Professor calculator? 
options: [To provide the answer to a mathematical expression, To generate unsolved expressions and prompt the user for the answer, To teach children how to use a regular calculator, To help children memorize multiplication tables.]
correct_answer: To generate unsolved expressions and prompt the user for the answer"
\end{minted}
    \caption{Grade 3 to 5 QA pair for computer science}
    \label{listing:3_5_pair_comp}
\end{listing}

\begin{listing}[!h]
\small
    \begin{minted}[frame=lines,
framesep=0.7mm,
baselinestretch=0.9,
fontsize=\footnotesize,
mathescape, 
breaklines]{bash}
"grade_level: grade 6 to 8
question: Why did Apple decide to change the case design of the first production Apple II computers, and what solution did they implement? 
answer: Apple decided to change the case design because the initial design had no vent openings, which caused high heat buildup from the PCB. This heat caused the plastic to soften and sag. To solve this problem, Apple added vent holes to the case within three months of production. Customers with the original case could have them replaced at no charge."
\end{minted}
    \caption{Grade 6 to 8 QA pair for computer science}
    \label{listing:6_8_pair_comp}
\end{listing}

\newpage

\begin{listing}[!h]
\small
    \begin{minted}[frame=lines,
framesep=0.7mm,
baselinestretch=1.1,
fontsize=\footnotesize,
mathescape, 
breaklines]{bash}
"grade_level: grade 9 to 12 
question: How did the development of mod_qos evolve to enhance the security of Apache servers over time?
answer: The development of mod_qos evolved to enhance the security of Apache servers by initially limiting the number of concurrent HTTP requests for specified resources. Over time, it introduced features such as connection level controls, request/response throttling, per client (IP) control mechanisms, and anomaly detection using client characteristic measurement. Additionally, it added geolocation features and improved response throttling, along with supporting differentiated services by controlling the DSCP IP field, all of which contributed to protecting servers against DoS attacks."
\end{minted}
    \caption{Grade 9 to 12 QA pair for computer science}
    \label{listing:9_12_pair_comp}
\end{listing}

\subsection{Medicine}

\begin{listing}[!h]
\small
    \begin{minted}[frame=lines,
framesep=0.7mm,
baselinestretch=1.1,
fontsize=\footnotesize,
mathescape, 
breaklines]{bash}
"grade_level: grade 1 and 2
question: What should you avoid to stay healthy? answer: Avoid touching people who are sick."
\end{minted}
    \caption{Grade 1 and 2 QA pair for medicine}
    \label{listing:1_2_pair_med}
\end{listing}

\begin{listing}[!h]
\small
    \begin{minted}[frame=lines,
framesep=0.7mm,
baselinestretch=1.1,
fontsize=\footnotesize,
mathescape, 
breaklines]{bash}
"grade_level: grade 3 to 5
question: Why is it important for people to get vaccinated against COVID-19? 
answer: Getting vaccinated helps protect people from getting sick with COVID-19. It also helps stop the virus from spreading to others, which can keep communities safer and help everyone return to normal activities more quickly."
\end{minted}
    \caption{Grade 3 to 5 QA pair for medicine}
    \label{listing:3_5_pair_med}
\end{listing}

\begin{listing}[!h]
\small
    \begin{minted}[frame=lines,
framesep=0.7mm,
baselinestretch=1.1,
fontsize=\footnotesize,
mathescape, 
breaklines]{bash}
"grade_level: grade 6 to 8
question: Why might people with severe COVID-19 need respiratory support, and what are some forms of this support?
answer: People with severe COVID-19 might need respiratory support because the virus can severely affect their lungs, making it difficult for them to breathe. Some forms of respiratory support include oxygen therapy, mechanical ventilation, and intravenous fluids."
\end{minted}
    \caption{Grade 6 to 8 QA pair for medicine}
    \label{listing:6_8_pair_med}
\end{listing}
\begin{listing}[!h]
    \begin{minted}[frame=lines,
framesep=0.7mm,
baselinestretch=1.1,
fontsize=\footnotesize,
mathescape, 
breaklines]{bash}
"grade_level: grade 9 to 12 
question: Analyze the impact of Dame Kate Isabel Campbell's discovery on the treatment of premature babies worldwide. How did her findings change medical practices? 
answer: Dame Kate Isabel Campbell's discovery that blindness in premature babies was caused by high concentrations of oxygen led to a significant change in medical practices worldwide. Her findings prompted a reevaluation of the treatment protocols for premature babies, specifically regarding the administration of oxygen. As a result, medical professionals adjusted the oxygen levels used in neonatal care to prevent blindness, thereby improving the health outcomes for premature infants globally."
\end{minted}
    \caption{Grade 9 to 12 QA pair for medicine}
    \label{listing:9_12_pair_med}
\end{listing}
\newpage
\subsection{Metrology}

\begin{listing}[!h]
\small
    \begin{minted}[frame=lines,
framesep=0.7mm,
baselinestretch=1.1,
fontsize=\footnotesize,
mathescape, 
breaklines]{bash}
"grade_level: grade 1 and 2
question: Can you name something that might be measured using a unit of volume?
answer: Water, rice, sugar, grain, or flour."
\end{minted}
    \caption{Grade 1 and 2 QA pair for metrology}
    \label{listing:1_2_pair_met}
\end{listing}

\begin{listing}[!h]
\small
    \begin{minted}[frame=lines,
framesep=0.7mm,
baselinestretch=1,
fontsize=\footnotesize,
mathescape, 
breaklines]{bash}
"grade_level: grade 3 to 5
question: Why do graduated cylinders have marked lines? answer: Graduated cylinders have marked lines to show the amount of liquid that has been measured. These lines help people accurately measure the volume of liquids in the cylinder."
\end{minted}
    \caption{Grade 3 to 5 QA pair for metrology}
    \label{listing:3_5_pair_met}
\end{listing}

\begin{listing}[!h]
\small
    \begin{minted}[frame=lines,
framesep=0.7mm,
baselinestretch=1,
fontsize=\footnotesize,
mathescape, 
breaklines]{bash}
"grade_level: grade 6 to 8
question: Explain how the volume of a cubic inch is related to a US gallon and why this might be useful in understanding volume conversions.
answer: A cubic inch is 1/231 of a US gallon. This relationship is useful for understanding volume conversions because it provides a way to translate between smaller units of volume (like cubic inches) and larger, more commonly used units (like gallons), which can be helpful in various practical applications such as cooking, fuel measurements, and fluid storage."
\end{minted}
    \caption{Grade 6 to 8 QA pair for metrology}
    \label{listing:6_8_pair_met}
\end{listing}
\begin{listing}[!h]
    \begin{minted}[frame=lines,
framesep=0.7mm,
baselinestretch=1,
fontsize=\footnotesize,
mathescape, 
breaklines]{bash}
"grade_level: grade 9 to 12 
question: How do enhanced geothermal systems (EGS) differ from traditional oil and gas fracking techniques in terms of environmental impact, and what measures are taken to minimize potential damage? 
answer: Enhanced geothermal systems (EGS) differ from traditional oil and gas fracking techniques primarily in their environmental impact. While both techniques involve injecting fluids under high pressure to expand rock fissures, EGS does not use toxic chemicals, reducing the possibility of environmental damage. Instead, EGS uses proppants like sand or ceramic particles to keep the cracks open and ensure optimal flow rates. Additionally, the geologic formations targeted by EGS are deeper, which further minimizes the risk of environmental harm."
\end{minted}
    \caption{Grade 9 to 12 QA pair for metrology}
    \label{listing:9_12_pair_met}
\end{listing}
\subsection{Ecology}

\begin{listing}[!h]
\small
    \begin{minted}[frame=lines,
framesep=0.7mm,
baselinestretch=0.9,
fontsize=\footnotesize,
mathescape, 
breaklines]{bash}
"grade_level: grade 1 and 2
question: What does the gecko mostly eat?,
answer: The gecko mostly eats insects."
\end{minted}
    \caption{Grade 1 and 2 QA pair for ecology}
    \label{listing:1_2_pair_eco}
\end{listing}
\begin{listing}[!h]
\small
    \begin{minted}[frame=lines,
framesep=0.7mm,
baselinestretch=0.9,
fontsize=\footnotesize,
mathescape, 
breaklines]{bash}
"grade_level: grade 3 to 5
question: Why do you think the white-winged swallow builds its nest a few meters above water?, 
answer: The white-winged swallow likely builds its nest a few meters above water to protect its eggs and young from predators and to ensure easy access to food, as swallows often feed on insects found near water."
\end{minted}
    \caption{Grade 3 to 5 QA pair for ecology}
    \label{listing:3_5_pair_eco}
\end{listing}

\begin{listing}[!h]
\small
    \begin{minted}[frame=lines,
framesep=0.7mm,
baselinestretch=1.1,
fontsize=\footnotesize,
mathescape, 
breaklines]{bash}
"grade_level: grade 6 to 8
question: What are the components included within the boundaries of the MPA, and why might it be important to protect all of these components? 
answer: The MPA includes the water column, the seabed, and the subsoil. Protecting all these components is important because they are interconnected ecosystems that support marine life, maintain biodiversity, and ensure the health of the marine environment. The water column provides habitat and resources for marine organisms, the seabed is home to various species and supports ecological processes, and the subsoil contains nutrients and minerals crucial for the overall ecosystem."
\end{minted}
    \caption{Grade 6 to 8 QA pair for ecology}
    \label{listing:6_8_pair_eco}
\end{listing}
\begin{listing}[!h]
    \begin{minted}[frame=lines,
framesep=0.7mm,
baselinestretch=1.2,
fontsize=\footnotesize,
mathescape, 
breaklines]{bash}
"grade_level: grade 9 to 12 
question: How does the long-eared myotis adapt its feeding strategy when hunting beetles, and why is this adaptation beneficial?
answer: The long-eared myotis adapts its feeding strategy by using its robust molars and highly placed articular process to exert more crushing force, allowing it to penetrate the hard carapace of beetles. This adaptation is beneficial because it enables the bat to effectively consume beetles, which have a protective shell that would otherwise be difficult to penetrate."
\end{minted}
    \caption{Grade 9 to 12 QA pair for ecology}
    \label{listing:9_12_pair_eco}
\end{listing}

\subsection{Geology}

\begin{listing}[!h]
\small
    \begin{minted}[frame=lines,
framesep=0.7mm,
baselinestretch=1.1,
fontsize=\footnotesize,
mathescape, 
breaklines]{bash}
"grade_level: grade 1 and 2
question: What does the Ely Limestone preserve?
answer: It preserves fossils"
\end{minted}
    \caption{Grade 1 and 2 QA pair for geology}
    \label{listing:1_2_pair_geology}
\end{listing}
\begin{listing}[!h]
\small
    \begin{minted}[frame=lines,
framesep=0.7mm,
baselinestretch=1.1,
fontsize=\footnotesize,
mathescape, 
breaklines]{bash}
"grade_level: grade 3 to 5
question: Why do aftershocks occur after a big earthquake?
answer: Aftershocks occur because the ground is still adjusting after the main earthquake. The big earthquake changes the stress in the Earth's crust, and the aftershocks are the Earth's way of settling back into a new balance."
\end{minted}
    \caption{Grade 3 to 5 QA pair for geology}
    \label{listing:3_5_pair_geology}
\end{listing}

\begin{listing}[!h]
\small
    \begin{minted}[frame=lines,
framesep=0.7mm,
baselinestretch=1.1,
fontsize=\footnotesize,
mathescape, 
breaklines]{bash}
"grade_level: grade 6 to 8
question: What processes contribute to the formation of floodplains, and how do they impact the landscape? answer: Floodplains are formed by the deposition of suspended load from overbank flow, bedload deposition from lateral river migration, and landscape processes such as landslides. These processes contribute to the buildup of land adjacent to river channels and shape the landscape by adding new layers of soil and altering the river's path."
\end{minted}
    \caption{Grade 6 to 8 QA pair for geology}
    \label{listing:6_8_pair_geology}
\end{listing}
\newpage
\begin{listing}[!h]
    \begin{minted}[frame=lines,
framesep=0.7mm,
baselinestretch=1.1,
fontsize=\footnotesize,
mathescape, 
breaklines]{bash}
"grade_level: grade 9 to 12 
question: Explain how erosional sheltering contributes to the formation of a crag and tail structure in rocks. answer: Erosional sheltering occurs when rocks contain particles that are harder than the surrounding material. As the rock is worn down, these harder particles resist erosion more effectively than the softer rock. This resistance protects the rock on the lee side of the hard particle from further wear. Over time, this process results in the formation of a crag, where the hard particle was located, and a tail that extends parallel to the direction of movement down-slip from the particle. The crag and tail structure is thus a direct result of the differential erosion rates between the hard particles and the surrounding softer rock."
\end{minted}
    \caption{Grade 9 to 12 QA pair for geology}
    \label{listing:9_12_pair_geology}
\end{listing}

\subsection{Geography}

\begin{listing}[!h]
\small
    \begin{minted}[frame=lines,
framesep=0.7mm,
baselinestretch=1,
fontsize=\footnotesize,
mathescape, 
breaklines]{bash}
"grade_level: grade 1 and 2
question: Is France in Europe?
answer: Yes, France is in Europe."
\end{minted}
    \caption{Grade 1 and 2 QA pair for geography}
    \label{listing:1_2_pair_geography}
\end{listing}
\begin{listing}[!h]
\small
    \begin{minted}[frame=lines,
framesep=0.7mm,
baselinestretch=1,
fontsize=\footnotesize,
mathescape, 
breaklines]{bash}
"grade_level: grade 3 to 5
question: Why might knowing the altitude and area of a municipality be important? 
answer: Knowing the altitude and area of a municipality can help us understand its climate, the types of plants and animals that live there, and how people might use the land. For example, higher altitudes might have cooler temperatures, and larger areas might have more space for homes, parks, or farms."
\end{minted}
    \caption{Grade 3 to 5 QA pair for geography}
    \label{listing:3_5_pair_geography}
\end{listing}
\begin{listing}[!h]
\small
    \begin{minted}[frame=lines,
framesep=0.7mm,
baselinestretch=1,
fontsize=\footnotesize,
mathescape, 
breaklines]{bash}
"grade_level: grade 6 to 8
question: Why is Île-de-France no longer considered an official wine region, and what recent developments suggest a revival in its viticulture? 
answer: Île-de-France is no longer considered an official wine region due to changes in wine region classifications. However, its viticulture is experiencing a revival, as evidenced by the establishment of more than 200 small recreational vineyards in recent decades, covering about 12 hectares, and the involvement of 5 villages in the Champagne area."
\end{minted}
    \caption{Grade 6 to 8 QA pair for geography}
    \label{listing:6_8_pair_geography}
\end{listing}
\begin{listing}[!h]
    \begin{minted}[frame=lines,
framesep=0.7mm,
baselinestretch=0.9,
fontsize=\footnotesize,
mathescape, 
breaklines]{bash}
"grade_level: grade 9 to 12 
question: Analyze how the historical significance of the Payap area has influenced its current role in Bangkok. Consider the changes in its name, infrastructure, and community over time. 
answer: The historical significance of the Payap area has greatly influenced its current role in Bangkok. Originally known as 'Payap,' meaning 'northwestern' or 'northern,' it was a significant location for the Chet Ton Dynasty, particularly as the residence of Prince Dilok Nopparat, son of King Chulalongkorn. This royal connection established Payap as a notable area in Bangkok. After the 1932 revolution, the area was renamed 'Si Yan,' meaning 'auspicious quarter,' reflecting a desire for good fortune and a new identity. Over time, the infrastructure evolved with the establishment of the Payap pier, which became a vital transport link via the Chao Phraya Express Boat. The community also transformed, becoming a diverse neighborhood with a mix of Thai and Chinese descent, reflecting its long history and resilience, especially during World War II when it housed a bomb shelter. Today, the area continues to serve as a functional and historical part of Bangkok, with the Ministry of Interior's Operation Centre for Displaced Persons and the Metropolitan Electricity Authority Hospital Samsen, indicating its ongoing importance in the city's social and administrative landscape."
\end{minted}
    \caption{Grade 9 to 12 QA pair for geography}
    \label{listing:9_12_pair_geography}
\end{listing}
\newpage
\section{Error Analysis of Model-Generated Responses}

\label{sec:AppendixF}

\begin{listing}[!h]
\small
\begin{minted}[frame=lines,
  framesep=0.7mm,
  baselinestretch=0.6,
  fontsize=\footnotesize,
  breaklines,
  autogobble]{json}
{ "type": "qa", "grade_level": "grade 1 and 2", "question": "What do we call a small outdoor space for horses that doesn't have a roof?", "answer": "A small outdoor space for horses without a roof is called a pen.", "model_answer": "A corral!  A corral is a small area outside a building where horses can graze and play. It has a fence or barrier around it to keep the horses in.", "Judge": {"Vocabulary Alignment": 6, "Conceptual Alignment": 5, "Scientific Language Alignment": 5, "Correctness": 4, "Clarity": 6, "Completeness": 4, "avg_rating": 5.0}}

{ "type": "qa", "grade_level": "grade 3 to 5", "question": "What is the purpose of built-in filters in some steam irons?", "answer": "Built-in filters in some steam irons are used to remove minerals from the water. This allows people to use standard tap water instead of distilled water, which helps prevent limescale build-up and can extend the lifespan of the iron.", "model_answer": "Sure, here's a clear and developmentally appropriate answer to the Built-in filters in some steam irons help to keep the iron clean and prevent rust. When you use an iron, the heat from the iron can cause the iron to become rusty over time. Built-in filters help to trap dust and other particles that would otherwise be carried into the iron and cause rust. This helps to keep your iron looking shiny and new.", "Judge": {"Vocabulary Alignment": 6, "Conceptual Alignment": 3, "Scientific Language Alignment": 4, "Correctness": 2, "Clarity": 6, "Completeness": 3, "avg_rating": 4.0}}

{ "type": "qa", "grade_level": "grade 6 to 8", "question": "Why might someone mistake bloating pain for heart pain, and where does this typically occur?", "answer": "Bloating pain might be mistaken for heart pain when it develops on the upper left side of the chest because the pain is sharp and can be severe.", "model_answer": "Bloating pain and heart pain are similar in that they both cause discomfort and can be quite painful. However, there are also some key differences between the two.
Bloating pain is often caused by gas, indigestion, or constipation. It can occur anywhere in the abdomen, but it is most common in the lower right abdomen.
Heart pain, on the other hand, is often caused by a heart condition, such as angina pectoris or pericarditis. It can also be caused by a muscle strain or injury to the chest wall. Heart pain is typically more severe and stabbing than bloating pain.
The pain from bloating and heart pain can be similar, so it can be difficult to tell the difference. However, if you are experiencing pain that is located in the lower right abdomen that is severe and lasts for more than a few minutes, it is important to see a doctor to get a diagnosis.", "Judge": {"Vocabulary Alignment": 6, "Conceptual Alignment": 3, "Scientific Language Alignment": 4, "Correctness": 2, "Clarity": 5, "Completeness": 3, "avg_rating": 3.8}}

{ "type": "qa", "grade_level": "grade 9 to 12", "question": "Explain how natural lasers differ from artificial lasers, and provide examples of celestial bodies where natural lasers have been observed.", "answer": "Natural lasers differ from artificial lasers in that they occur naturally in space without human intervention, whereas artificial lasers are man-made devices that require specific conditions and materials to function. Natural lasers are produced when irradiated planetary or stellar gases amplify light, similar to astrophysical masers. Examples of celestial bodies where natural lasers have been observed include Mars, Venus, and the star MWC 349.", "model_answer": "## Natural vs. Artificial Lasers
**Natural Lasers:**
* **Origin in the Earth's atmosphere:** Natural lasers are generated deep within the Earth's atmosphere, typically in the form of fluorescence or stimulated emission.
* **Do not require external energy:** They are powered by the energy of the sun, which excites atoms and molecules to emit light.
* **Examples:** The most famous natural laser is the **Green Flash** in the night sky, caused by the interaction of sunlight with water droplets in the atmosphere.
* **Do not require a medium to operate:** They can operate in a vacuum or a transparent medium like water vapor.
**Artificial Lasers:**
* **Man-made devices:** They are created by humans and do not occur naturally in the environment.
* **External energy source required:** They require an external energy source, such as electricity, to excite the active medium.
* **Examples:** Some common artificial lasers include **Lasers, LEDs, and fiber-optic communication systems**.
* **Need a medium to operate:** They require a medium, such as a gas, liquid, or solid, to interact with.
## Celestial Bodies with Natural Lasers
* **Sun:** The Sun's corona, a hot, high-temperature plasma, is a natural laser that illuminates space with its green light.
* **Stars:** Stars generate light through nuclear fusion in their cores, creating light and heat.
* **Comets:** Some comets and asteroids exhibit periodic variations in brightness due to the presence of ice and dust particles that act as natural lasers.
* **Lasers in space:** Natural lasers are also found in space, such as in the Aurora Borealis and the red giant star Vega.", "Judge": {"Vocabulary Alignment": 6, "Conceptual Alignment": 3, "Scientific Language Alignment": 5, "Correctness": 2, "Clarity": 5, "Completeness": 3, "avg_rating": 4.0}}
\end{minted}
\caption{Gemma-2B Model Generated Responses}
\label{listing:gemma}
\end{listing}

\begin{listing}[!h]
\small
\begin{minted}[frame=lines,
  framesep=0.7mm,
  baselinestretch=0.6,
  fontsize=\footnotesize,
  mathescape,
  breaklines,
  autogobble]{json}
{ "type": "qa", "grade_level": "grade 1 and 2", "question": "What kind of water does a direct piping water dispenser give?", "answer": "It gives hot or cold filtered pure drinking water.", "model_answer": "Yay! Let me tell you a secret! A direct piping water dispenser gives... COLD WATER! It's like a special machine that brings fresh, cool water right to your table! Isn't that cool?", "Judge": {"Vocabulary Alignment": 7, "Conceptual Alignment": 3, "Scientific Language Alignment": 3, "Correctness": 2, "Clarity": 6, "Completeness": 2, "avg_rating": 3.8}}

{ "type": "qa", "grade_level": "grade 3 to 5", "question": "What kind of areas does the ashy flowerpecker like to live in?", "answer": "The ashy flowerpecker likes to live in forests and more degraded woody areas like gardens and farm edges.", "model_answer": "What a great question! The ashy flowerpecker is a type of bird that loves to live in special places. It likes to live in areas with lots of trees, like forests or woodlands. It also likes to live near water, like rivers, lakes, or even wetlands. The ashy flowerpecker likes to build its nest in the branches of trees, usually high up where it's safe and cozy. Imagine you're on a nature walk, and you're surrounded by tall trees and the sound of birds chirping. That's where the ashy flowerpecker might be living!", "Judge": {"Vocabulary Alignment": 7, "Conceptual Alignment": 4, "Scientific Language Alignment": 4, "Correctness": 3, "Clarity": 6, "Completeness": 3, "avg_rating": 4.5}}

{ "type": "qa", "grade_level": "grade 6 to 8", "question": "What percentage of the ecoregion is protected, and why might protecting such a large area be important for the environment?", "answer": "60.3% of the ecoregion is protected. Protecting such a large area is important for the environment because it helps preserve biodiversity, maintain natural habitats for wildlife, and protect ecosystems from human activities that could cause harm, such as pollution or deforestation. It also helps in maintaining ecological balance and ensuring that natural resources are available for future generations.", "model_answer": "Let's dive into the world of conservation! The ecoregion we're talking about is the Amazon rainforest. According to recent studies, approximately 20% of the Amazon rainforest is protected. This means that about one-fifth of the entire region is safeguarded by national parks, indigenous reserves, and other protected areas. Protecting such a large area is crucial for the environment for several reasons: **Biodiversity hotspot**: The Amazon rainforest is home to an incredible array of plant and animal species, many of which can be found nowhere else on Earth. By protecting this area, we're preserving the habitats of these unique creatures and ensuring their survival. **Carbon sequestration**: The Amazon rainforest plays a significant role in absorbing carbon dioxide from the atmosphere, which helps to combat climate change. By preserving the forest, we're helping to reduce the amount of greenhouse gases in the atmosphere. **Soil conservation**: The Amazon rainforest's dense vegetation helps to hold the soil in place, preventing erosion and landslides. This is especially important in areas prone to heavy rainfall and flooding. **Indigenous communities**: Many indigenous communities rely on the Amazon rainforest for their livelihoods, culture, and traditional way of life. Protecting the forest helps to preserve their rights and way of life. **Ecosystem services**: The Amazon rainforest provides numerous ecosystem services, such as clean air and water, that benefit not only the local population but also the global community. In summary, protecting 20% of the Amazon rainforest is essential for maintaining the region's biodiversity, combating climate change, conserving soil, supporting indigenous communities, and providing essential ecosystem services.", "Judge": {"Vocabulary Alignment": 7, "Conceptual Alignment": 4, "Scientific Language Alignment": 5, "Correctness": 2, "Clarity": 6, "Completeness": 3, "avg_rating": 4.5}}

{ "type": "qa", "grade_level": "grade 9 to 12", "question": "Why is the phase curve of Earth not determined as accurately as those for Mercury and Venus, and what method has been used as a proxy to study it?", "answer": "The phase curve of Earth is not determined as accurately as those for Mercury and Venus because its integrated brightness is difficult to measure from the surface. Instead, earthshine reflected from the portion of the Moon not lit by the Sun has been used as a proxy to study it.", "model_answer": "The phase curve of Earth is not determined as accurately as those for Mercury and Venus because it's difficult to observe the phase changes of our own planet from space. This is because Earth is the third planet from the Sun, and its proximity to the Sun makes it challenging to capture high-quality images of its phases from space. Unlike Mercury and Venus, which are closer to the Sun and have a more extreme tilt in their orbits, Earth's orbit is relatively stable and its distance from the Sun is not extreme. This makes it harder to capture clear images of its phases from space. To study the phase curve of Earth, scientists have used a proxy method called \"transit photometry.\" This involves measuring the decrease in brightness of a star as Earth passes in front of it, creating a mini-eclipse. By analyzing the amount of light blocked by Earth, scientists can infer its size and shape, which helps them estimate its phase curve. In this method, astronomers use space-based telescopes like the Kepler Space Telescope and the Transiting Exoplanet Survey Satellite (TESS) to monitor the brightness of stars and detect the tiny changes caused by Earth's passage. By analyzing these data, scientists can reconstruct the phase curve of our planet, providing valuable insights into its size, shape, and atmospheric properties. So, while we can't directly observe the phase curve of Earth from space, scientists have developed innovative methods like transit photometry to study our planet's phases and gain a better understanding of its characteristics.", "Judge": {"Vocabulary Alignment": 7, "Conceptual Alignment": 3, "Scientific Language Alignment": 5, "Correctness": 2, "Clarity": 6, "Completeness": 3, "avg_rating": 4.3}}
\end{minted}
\caption{Llama3-8B Model Generated Responses}
\label{listing:llama}
\end{listing}

\begin{listing}[!h]
\small
\begin{minted}[frame=lines,
  framesep=0.7mm,
  baselinestretch=0.6,
  fontsize=\footnotesize,
  mathescape,
  breaklines,
  autogobble]{json}
{ "type": "qa", "grade_level": "grade 1 and 2", "question": "What should you avoid to stay healthy?", "answer": "Avoid touching people who are sick.", "model_answer": "To stay healthy, you should avoid a few things:  1. **Eating too much junk food**: Like candy, chips, and soda. They're okay sometimes, but too much can make you feel sick.  2. **Not washing your hands**: Always wash your hands with soap and water, especially before eating and after using the bathroom. This helps stop germs from making you sick.  3. **Not getting enough sleep**: Your body needs rest to grow and stay strong. Try to sleep for 9-12 hours every night.  4. **Not exercising**: Play and move around every day. It helps your body stay strong and healthy.  5. **Not listening to your body**: If you feel sick or hurt, tell a grown-up. They can help you feel better.", "Judge": {"Vocabulary Alignment": 7, "Conceptual Alignment": 4, "Scientific Language Alignment": 4, "Correctness": 3, "Clarity": 6, "Completeness": 3, "avg_rating": 4.5}}

{ "type": "qa", "grade_level": "grade 3 to 5", "question": "Besides small mammals, what other types of animals does the unspotted saw-whet owl eat?", "answer": "The unspotted saw-whet owl is also thought to eat birds and bats.", "model_answer": "The unspotted saw-whet owl is a small owl that eats a variety of animals besides small mammals. Some of the other types of animals it eats include:  1. **Insects**: Like beetles, moths, and other bugs. 2. **Birds**: Sometimes it eats other small birds. 3. **Reptiles and Amphibians**: Such as small snakes, lizards, and frogs. 4. **Other small animals**: Like earthworms and even small fish if they live near water.  So, the unspotted saw-whet owl has a pretty varied diet!", "Judge": {"Vocabulary Alignment": 7, "Conceptual Alignment": 3, "Scientific Language Alignment": 4, "Correctness": 2, "Clarity": 6, "Completeness": 3, "avg_rating": 4.2}}

{ "type": "qa", "grade_level": "grade 6 to 8", "question": "Why is Good Technology important for businesses, and which industries benefit from its services?", "answer": "Good Technology is important for businesses because it provides mobile security solutions that help manage and secure mobile devices, apps, and data in a business environment. Industries that benefit from its services include financial services, healthcare, manufacturing, energy and utilities, legal, government, and technology.", "model_answer": "Good technology is super important for businesses because it helps them work more efficiently, communicate better, and stay competitive. Here are a few reasons why:\n1. **Efficiency**: Technology automates tasks that would otherwise take a lot of time to do by hand. For example, computers can process large amounts of data quickly, and software can handle repetitive tasks like sending emails or managing inventory.\n2. **Communication**: With tools like email, video conferencing, and instant messaging, businesses can communicate with customers, partners, and employees all over the world quickly and easily.\n3. **Data Management**: Technology helps businesses store, organize, and analyze data. This can help them make better decisions, understand their customers better, and track their progress.\n4. **Innovation**: Technology allows businesses to create new products and services, which can help them stay ahead of the competition.\nMany industries benefit from good technology. Here are a few examples:\n1. **Healthcare**: Technology helps doctors and nurses manage patient records, communicate with each other, and even perform surgeries with robotic assistance.\n2. **Education**: Schools use technology for online learning platforms, digital textbooks, and educational apps that make learning more interactive and accessible.\n3. **Retail**: Online shopping, inventory management systems, and customer relationship management (CRM) software help retailers manage their stores and serve customers better.\n4. **Transportation**: GPS systems, route planning software, and even self-driving technology help transportation companies operate more efficiently and safely.\n5. **Entertainment**: Streaming services, video games, and social media platforms rely heavily on technology to provide entertainment and connect people.\nIn short, good technology is like a superpower for businesses, helping them to do more, do it better, and do it faster!", "Judge": {"Vocabulary Alignment": 7, "Conceptual Alignment": 4, "Scientific Language Alignment": 5, "Correctness": 3, "Clarity": 7, "Completeness": 4, "avg_rating": 5.0}}

{ "type": "qa", "grade_level": "grade 9 to 12", "question": "How does Flammer syndrome primarily affect the body, and what are some of its manifestations?", "answer": "Flammer syndrome primarily affects the body through dysregulation of the blood supply, leading to symptoms such as cold hands and feet, and is often associated with low blood pressure.", "model_answer": "Flammer syndrome is a condition that primarily affects the blood vessels in the body, particularly those in the eyes and the heart. It's often associated with a tendency for the blood vessels to constrict (narrow) more than usual. This can lead to several different symptoms and issues.\nHere are some of the main ways Flammer syndrome can affect the body and its manifestations:\n1. **Eye Problems**:\n   - **Glaucoma**: Flammer syndrome can increase the risk of developing glaucoma, a condition where the pressure inside the eye increases, which can damage the optic nerve.\n   - **Retinal Vein Occlusion**: This is a condition where a vein in the retina becomes blocked, which can lead to vision problems.\n   - **Dry Eyes**: Some people with Flammer syndrome may experience dry eyes due to reduced blood flow to the eye.\n2. **Heart and Circulation Issues**:\n   - **Raynaud's Phenomenon**: This is a condition where the blood vessels in the fingers and toes constrict in response to cold or stress, causing them to turn white or blue and feel numb or painful.\n   - **High Blood Pressure**: Flammer syndrome can sometimes be associated with high blood pressure, especially in the eyes.\n3. **Other Symptoms**:\n   - **Cold Hands and Feet**: Due to reduced blood flow, people with Flammer syndrome may often have cold hands and feet.\n   - **Migraines**: Some people with Flammer syndrome may experience migraines, which are severe headaches often accompanied by nausea and sensitivity to light and sound.\n4. **General Health**:\n   - **Fatigue**: Reduced blood flow can sometimes lead to feelings of tiredness or fatigue.\n   - **Sleep Issues**: Some people with Flammer syndrome may have trouble sleeping, which can be related to the body's response to stress and blood flow issues.", "Judge": {"Vocabulary Alignment": 8, "Conceptual Alignment": 6, "Scientific Language Alignment": 7, "Correctness": 5, "Clarity": 8, "Completeness": 6, "avg_rating": 6.7}}
\end{minted}
\caption{Mistral-Small-24B Model Generated Responses}
\label{listing:mistral}
\end{listing}

\end{document}